\documentclass[10pt,twocolumn,letterpaper]{article}

\usepackage[pagenumbers]{iccv} 

\usepackage[accsupp]{axessibility}
\usepackage[table]{xcolor}
\usepackage{tabularx}
\usepackage{graphicx}
\usepackage{multirow}
\usepackage{algorithm, algorithmic}
\usepackage[resetlabels]{multibib}
\usepackage{marvosym}

\definecolor{color1}{HTML}{FFE4D6}
\definecolor{color2}{HTML}{ADD8E6}

\definecolor{iccvblue}{rgb}{0.21,0.49,0.74}
\usepackage[pagebackref,breaklinks,colorlinks,allcolors=iccvblue]{hyperref}

\makeatletter
\def\customsymbol#1{
    \ifcase\number\value{#1}
        \or *
        \or \Letter
        \or 1
    \else\@ctrerr
    \fi
}
\makeatother

\def\paperID{7230} 
\def\confName{ICCV}
\def\confYear{2025}

\title{When Confidence Fails: Revisiting Pseudo-Label Selection in \\
Semi-supervised Semantic Segmentation}

\author{Pan Liu, Jinshi Liu\footnotemark[1]\\
Central South University \\
{\tt\small \{pan.liu, ljs11528\}@csu.edu.cn}
}

\begin{document}
\maketitle
\setcounter{footnote}{1}
\renewcommand{\thefootnote}{\customsymbol{footnote}}
\footnotetext[1]{Corresponding author.} 

\begin{abstract}
While significant advances exist in pseudo-label generation for semi-supervised semantic segmentation, pseudo-label selection remains understudied. Existing methods typically use fixed confidence thresholds to retain high-confidence predictions as pseudo-labels. However, these methods cannot cope with network overconfidence tendency, where correct and incorrect predictions overlap significantly in high-confidence regions, making separation challenging and amplifying model cognitive bias. Meanwhile, the direct discarding of low-confidence predictions disrupts spatial-semantic continuity, causing critical context loss. We propose Confidence Separable Learning (CSL) to address these limitations. CSL formulates pseudo-label selection as a convex optimization problem within the confidence distribution feature space, establishing sample-specific decision boundaries to distinguish reliable from unreliable predictions. Additionally, CSL introduces random masking of reliable pixels to guide the network in learning contextual relationships from low-reliability regions, thereby mitigating the adverse effects of discarding uncertain predictions. Extensive experimental results on the Pascal, Cityscapes, and COCO benchmarks show that CSL performs favorably against state-of-the-art methods. Code and model weights are available at:\href{https://github.com/PanLiuCSU/CSL}{\textit{\texttt{https://github.com/PanLiuCSU/CSL}}}.
\end{abstract}
    
\section{Introduction}
\label{sec:intro}

\begin{figure}[t]
  \centering
   \includegraphics[width=0.748\linewidth]{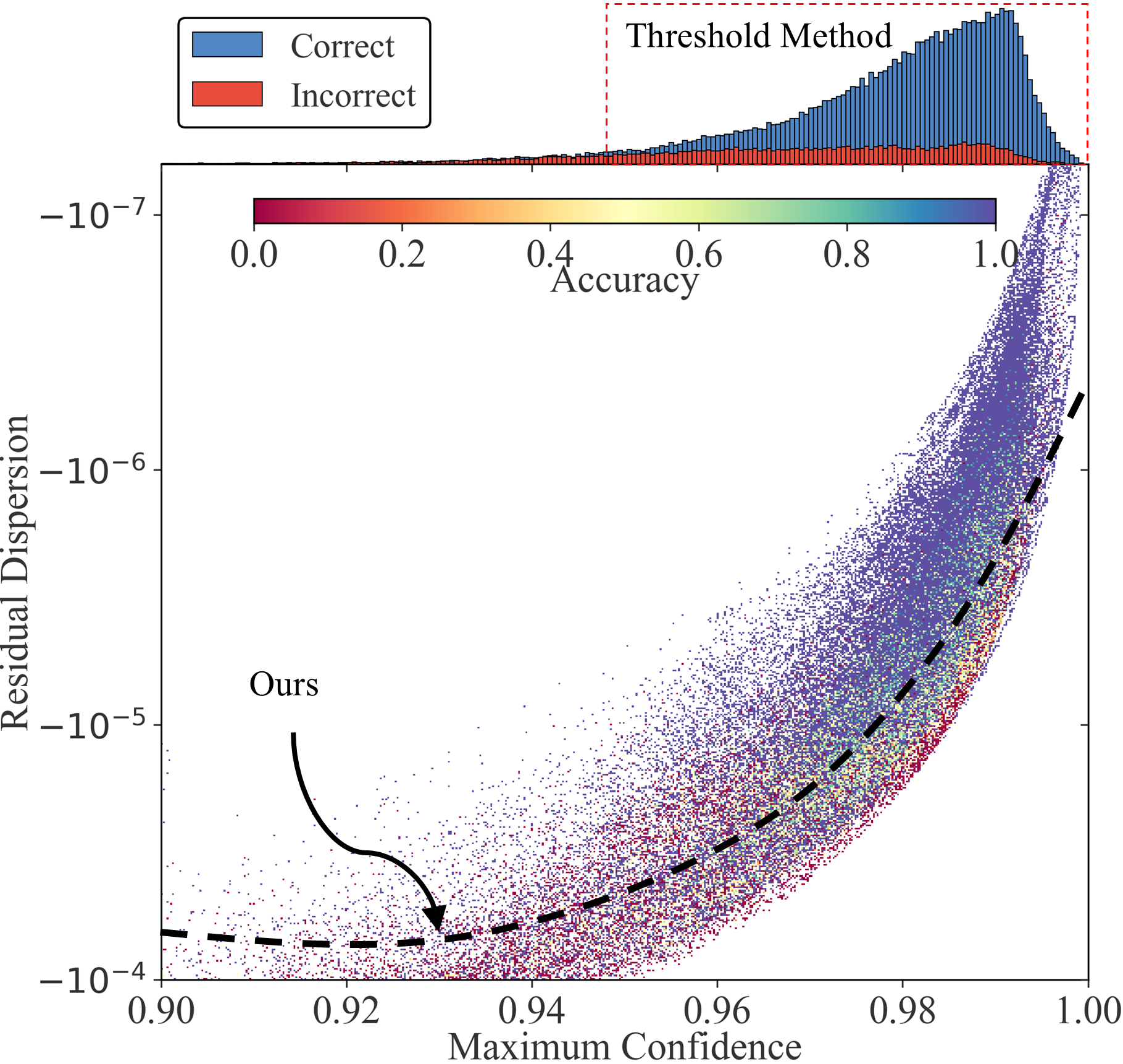}
   \caption{ prediction in confidence feature space. Blue/red points and upper histogram denote correct/incorrect predictions. The threshold method (dashed box) poorly separates them, while our approach (black curve) achieves clear discrimination.} 
   \label{fig:1}
\end{figure}

Semantic segmentation is a dense prediction task involving per-pixel classification. This technique is widely applied in critical domains such as autonomous driving and medical imaging. However, fully supervised segmentation methods ~\cite{deeplab,ljs2,Seg1} often require extensive high-quality, pixel-level labeled datasets, which can be costly and resource-intensive. In light of these challenges, semi-supervised semantic segmentation ~\cite{Decoupled,Seg4,Seg2} has emerged as a promising research area. This approach seeks to leverage a limited number of labeled images alongside a substantial quantity of unlabeled images to improve segmentation performance.

In the field of semi-supervised semantic segmentation, combining pseudo-labeling with consistency regularization has become a popular strategy~\cite{mittal2019semi}. Pseudo-labeling methods~\cite{RCPS,ljs1, cps,three_self_training} generate pseudo-labels for unlabeled samples based on the model predictions to optimize the model iteratively. Consistency regularization methods~\cite{CutMix,CAC,rankmatch,mittal2019semi} follow the smoothness assumption~\cite{survey}, encouraging the model to generate consistent predictions under different perturbations of the same sample. The effectiveness of these strategies depends heavily on the quality of the pseudo-labels, and existing methods typically assume a direct correlation between high confidence and high accuracy, relying on the maximum confidence scores to filter or weight the pseudo-labels~\cite{fixmatch, DAW, AEL}.

As shown by the red box in \cref{fig:1}, neural networks often suffer from overconfidence, causing predictions to cluster at extreme confidence levels regardless of their correctness~\cite{calibration}. This issue becomes more pronounced when labeled data is scarce, as the limited annotations are insufficient to effectively calibrate the model's confidence estimates on the vast unlabeled dataset. Such overconfidence presents significant challenges for semi-supervised semantic segmentation. Firstly, it renders confidence scores poor indicators of prediction accuracy, since both correct and incorrect predictions may receive similarly high confidence scores due to miscalibration. Secondly, constrained by the low distinguishability of confidence, existing methods are forced to exclude low-confidence predictions to avoid introducing errors during training~\cite{bias}, but this also results in a lack of supervision in borderline or complex regions of the data, that are typically crucial for model improvement. 

We propose the \textbf{Confidence Separable Learning} (CSL) framework to address the challenge of insufficient discriminability of pseudo-labels. Based on entropy minimization theory, we find that reliable prediction should have both high maximum confidence and significant probability dispersion between non-maximum class probabilities. Therefore, we construct a confidence distribution feature space consisting of maximum confidence and residual deviation to guide the selection process. As shown in \cref{fig:1}, while an overconfident network assigns similar confidence scores to correct and incorrect predictions, these two classes of predictions exhibit a high degree of separability in the confidence distribution feature space. Based on this property, CSL adaptively establishes sample-specific pseudo-label decision boundaries by formulating pseudo-label selection as a convex optimization problem that maximizes inter-class separability. This allows CSL to more adequately extract supervised signals from unlabelled data.

While CSL enhances pseudo-label reliability, its inherent precision-coverage trade-off results in the persistent exclusion of low-reliability regions from supervision, potentially amplifying regional supervision bias. To address this, we integrate a Trusted Mask Perturbation (TMP) strategy into our CSL framework. By randomly masking subsets of reliable predictions, we force the model to reconstruct semantics from adjacent unreliable regions via contextual dependencies. This process maximizes mutual information between unreliable regions and pseudo-labeled areas, thereby mitigating regional bias induced by selective pseudo-labeling. Our contributions can be summarized as:

\begin{itemize}
\item[$\bullet$] We demonstrate that pseudo-label selection based only on maximum confidence suffers from suboptimal performance, especially when the network is overconfident.
\item[$\bullet$] We propose a semi-supervised semantic segmentation framework that separates reliable predictions by maximizing separability optimization strategy and constructs effective supervisory signals for unreliable predictions.
\item[$\bullet$] Comprehensive experiments on three benchmark datasets confirm the effectiveness of our proposed method.
\end{itemize}
\section{Related work}
\label{sec:Related}

\textbf{Semi-supervised learning (SSL)}~\cite{Unsupervised,kim2022conmatch, Usb} aims to enhance model performance by leveraging labeled and unlabeled data.  Recent investigations have concentrated on the techniques of self-training
~\cite{chen2022debiased,xie2020self,zoph2020rethinking,vat,FST} and consistency regularization~\cite{MT,xu2022semi,Temporal,sregularization,fixmatch}.  However, these methods limited by marginal distribution discrepancies, leading to significant erroneous supervision.

To address this limitation, two main research directions have been explored. The first focuses on optimizing threshold-based pseudo-labeling strategies~\cite{Dash,softmatch,freematch,flexmatch}. SoftMatch~\cite{softmatch} models confidence as a Gaussian distribution to determine optimal thresholds. Despite improvements, these methods rely on assumptions that may not be universally applicable, limiting their generalizability. The second seeks additional metrics beyond confidence, such as Monte Carlo dropout~\cite{defense}, ensemble methods~\cite{Double}, and auxiliary discriminators~\cite{SemiReward}. However, the additional metrics significantly increase the computational overhead.

\noindent\textbf{Semi-supervised semantic segmentation} faces the challenge of generating accurate pixel-level supervisory signals for unlabeled images. Inspired by SSL, many works have adopted similar frameworks ~\cite{Seg3,Unimatchv2,allspark}. PS-MT~\cite{PS-MT} uses dual-teacher models to enhance consistency regularization. UniMatch~\cite{UniMatch} separates feature augmentation from image augmentation to reduce learning difficulty. Augmentation techniques like Cowmix~\cite{cowmix} focus on providing diverse training samples.

However, these methods generally do not directly improve the quality of pseudo-labels. To overcome these limitations, several approaches have been proposed~\cite{DARS,ELN}. AEL~\cite{AEL} employs a sample weighting strategy to emphasize reliable predictions. DAW~\cite{DAW} uses labeled data distribution to determine optimal thresholds. LOGICDIAG~\cite{LOGICDIAG} introduces symbolic reasoning to correct erroneous pseudo-labels. 

Nevertheless, these methods are still constrained by the low separability of confidence scores, resulting in biased supervisory signals. Therefore, ESL ~\cite{ESL} uses multi-hot labels to mitigate the effects of incorrect predictions. U2PL~\cite{U2PL} leverages unreliable predictions through contrastive learning. CorrMatch~\cite{CorrMatch} utilizes correlation maps to propagate pseudo-labels in low-confidence predictions. Furthermore, this pseudo-label selection challenge exists even in unsupervised adaptation. TransCal~\cite{PC} introduces calibration loss to improve the calibration performance of labelless sets. MIC~\cite{MIC} employing random masking strategies attempt to enhance prediction reliability. Despite these efforts, these methods face difficulties in fully eliminating inherent biases. Unlike previous approaches, CSL introduces additional metrics to address the low discriminability of confidence scores and explicitly mitigates the supervision bias caused by pseudo-labels.
\begin{figure*}[t]
  \centering
   \includegraphics[width=1\linewidth]{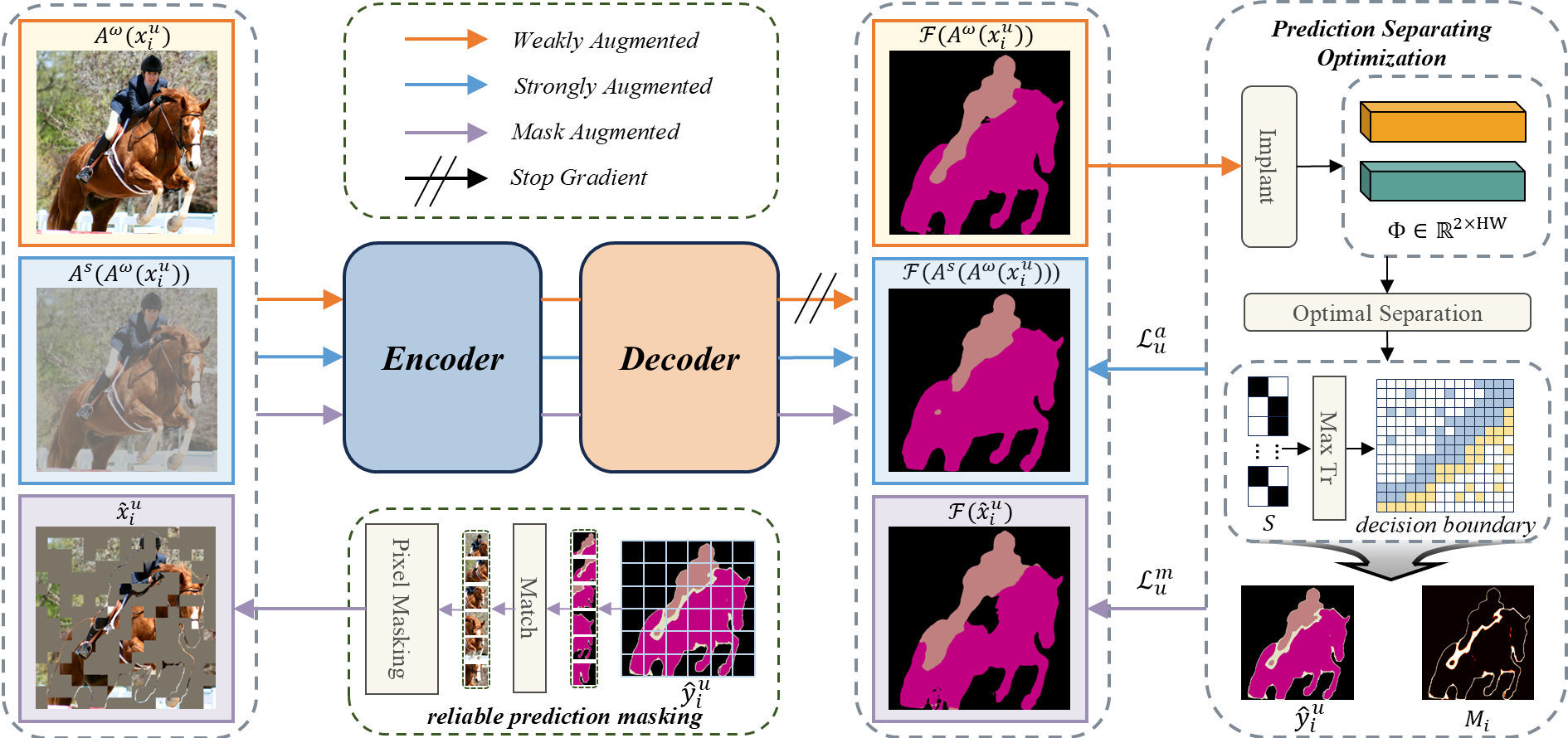}
   \caption{Overview of CSL pipeline for unlabeled images. The predictions of weakly augmented images are partitioned on a per-pixel basis into reliable and unreliable predictions through a convex optimization strategy in the feature space $\Phi$. This process constructs loss weight mask $M_i$ and pseudo-label $\hat{y}_i^u$, which are then used as supervisory signals for strongly augmented and masked images.} 
   \label{fig:2}
\end{figure*}

\section{Method}
\label{sec:method}
\subsection{Preliminaries}
\label{subsec:3.1}

In semi-supervised semantic segmentation, the training data consists of a labeled set \( D_l = \{(x_i^l, y_i^l)\}_{i=1}^{N_l} \) and an unlabeled set \( D_u = \{x_i^u\}_{i=1}^{N_u} \), where \( N_l \ll N_u \). \( x_i \in \mathbb{R}^{3 \times H \times W} \) is the input image, with $H$ and $W$ denoting its height and width, respectively. \( y_i \in \mathbb{R}^{K \times H \times W} \) is the one-hot encoded pixel label map with \( K \) classes. For the semantic segmentation network \( \mathcal{F} \), the supervised loss is defined as:

\begin{equation}
    \mathcal{L}_S = \frac{1}{B_L} \sum_{i=1}^{B_L} CE\left(y_i^l, \mathcal{F}\left(A^\omega\left(x_i^l\right)\right)\right),
    \label{eq:1}
\end{equation}
where \( B_L \) is the batch size of labeled data, \( CE(\cdot, \cdot) \) denotes the cross-entropy loss function, and \( A^\omega(\cdot) \) is the weak data augmentation function.

Similarly, given a batch size \( B_U \) of unlabeled data, the unsupervised training loss follows a simple framework ~\cite{UniMatch} with weak-to-strong consistency regularization:
\begin{equation}
    \mathcal{L}_u^a = \frac{1}{B_U} \sum_{i=1}^{B_U} M_i \odot CE\left(\hat{y}_i^u, \mathcal{F}\left(A^s\left(A^\omega\left(x_i^u\right)\right)\right)\right),
    \label{eq:2}
\end{equation}
where \( \hat{y}_i^u \) is the label derived from \( F\left(A^\omega\left(x_i^u\right)\right) \), \( A^s(\cdot) \) is the strong data augmentation function, and \( M_i \) is the indicator mask for pseudo-labels in \( \hat{y}_i^u \). In the baseline, following the confidence thresholding method,
\begin{equation}
    M_i = \mathbb{I}\left(\max\left(F\left(A^\omega\left(x_i^u\right)\right)\right) > \tau\right),
    \label{eq:3}
\end{equation}
where \( \mathbb{I}(\cdot) \) is the indicator function, \( \tau \) denotes the confidence threshold. However, this strategy inevitably suffers from cognitive bias and supervision loss.

To address these issues, we propose CSL. As illustrated in \cref{fig:2}, CSL introduces residual dispersion as an additional measure(\cref{subsec:3.2}) and through spectral relaxation, formulates $M_i$ construction as a convex optimization problem(\cref{subsec:3.3}). Subsequently, CSL introduces additional supervisory signals that compel the network to learn global context relationships based on regions lacking supervision, compensating for supervisory gaps in challenging areas (\cref{subsec:3.4}).

\subsection{Theoretical Analysis of Residual Dispersion}
\label{subsec:3.2}

The overconfidence tendency of deep networks challenges the effectiveness of maximum confidence as the sole criterion for pseudo-label selection. Based on the entropy minimization principle, we theoretically demonstrate that reliable predictions should simultaneously exhibit high maximum confidence and residual dispersion.

Let $p_n \in\mathbb{R}^K$ denote the predicted class probabilities for pixel $n$ in an unlabeled image $x_i^u$, \( p_n(k') = \max_k p_n(k) \) represents the maximum confidence. According to entropy minimization, the ideal prediction should follow an unimodal distribution \( q = [q_1,...,q_K] \), where \(q_{k'} = 1 - (K-1)\epsilon\) and \(q_{k \neq k'} = \epsilon\) with \(\epsilon \to 0\). Based on the decomposition form of cross-entropy, the goal can be defined as
\begin{equation}
CE(p_n, q) = -q_{k'}\log p_n(k') - \epsilon \sum_{k \neq k'} \log p_n(k).
\label{eq:entropy_decomp}
\end{equation}
Decomposing non-maximum probabilities $p_n(k),\forall k  \neq k'$ into mean-shift components $p_n(k) = \mu_{\text{res}} + \delta_k$, where \(\mu_{\text{res}} = \frac{1-p_n(k')}{K-1}\) and $\sum\nolimits_{k \neq k'} \delta_k = 0$.
Applying second-order Taylor expansion to \(\log p_n(k)\) around \(\mu_{\text{res}}\),
\begin{equation}
\log p_n(k) = \log \mu_{\text{res}} + \frac{\delta_k}{\mu_{\text{res}}} - \frac{\delta_k^2}{2\mu_{\text{res}}^2} + \mathcal{O}(\delta_k^3), \forall k  \neq k'.
\label{eq:taylor_expansion}
\end{equation}
Substituting \cref{eq:taylor_expansion} into \cref{eq:entropy_decomp} and simplifying,
\begin{align}
CE(p_n, q) \approx -\log p_n(k') - \frac{\epsilon(K-1)^3}{2(1-p_n(k'))^2} \cdot v_n,
\label{eq:v_derivation}
\end{align}
where the \textbf{residual dispersion} \(v_n\) is defined as
\begin{equation}
v_n \triangleq -\frac{1}{K-1}\sum_{k \neq k'}\delta_k^2.
\label{eq:v_def}
\end{equation}\

\cref{eq:v_derivation} reveals that under entropy minimization objectives, trustworthy predictions require maximizing both maximum confidence $p_n(k')$ and residual dispersion $v_n$. Notably, as $p_n(k') \to 1$, the coefficient term of $v_n$ becomes significant, shows that $v_n$ becomes a key indicator to distinguish prediction reliability in overconfidence cases.

\subsection{Prediction Convex Optimization Separation}
\label{subsec:3.3}

\cref{subsec:3.2} shows that prediction reliability nonlinearly depends on both $ p_n(k') $ and $ v_n $. However, threshold-based methods fail to capture the nonlinear complementary relationship. To overcome this limitation, we propose a sample-adaptive method, which formulates pseudo-label selection as a convex optimization problem by spectral relaxation.

We construct an attribute vector $h_n=[p_{n}(k'), v_n]^T$ for each pixel prediction, embedding it into a feature space that captures maximum confidence and residual dispersion. Based on this, we aim to separate predictions with high maximum confidence and residual dispersion to select reliable pseudo-labels. This selection is formalized as a trace maximization problem,
\begin{equation}
    \mathop{\max}\limits_{\mathbf{S}}Tr\left(S^T \Phi^T \Phi S\right), s.t.S\in\{0,1\}^{HW\times2},
    \label{eq:5}
\end{equation}
where $Tr(\cdot)$ denotes the trace of a matrix. $\Phi=[\mathbf{h}_1,\mathbf{h}_2,...,\mathbf{h}_{HW}]$ is the feature matrix of $x_i^u$. $S$ is the selection matrix, subject to the constraint $\sum_{c} S_{n,c} = 1$, with $c \in \{1,2\}$ indicates the class index in the selection.

To solve this non-convex combinatorial optimization problem, we apply spectral relaxation~\cite{relaxation}, relaxing the discrete constraints on $S$. According to the Ky Fan theorem ~\cite{kf}, the optimal solution $S^\ast$ approximates:
\begin{equation}
    S_{n,c}^\ast = \Delta\left(\arg\max_{i \in \{1,2\}} |u_{i}(n)|, c\right),
    \label{eq:6}
\end{equation}
where $u_1,u_2$ are the eigenvectors of the matrix $\Phi^T \Phi$, $|u_{i}(n)|$ denotes the absolute value of the $n$-th element in $u_{i}$, and $\Delta(\cdot,\cdot)$ is the Kronecker delta function.

\begin{figure}[t]
  \centering
   \includegraphics[width=0.9\linewidth]{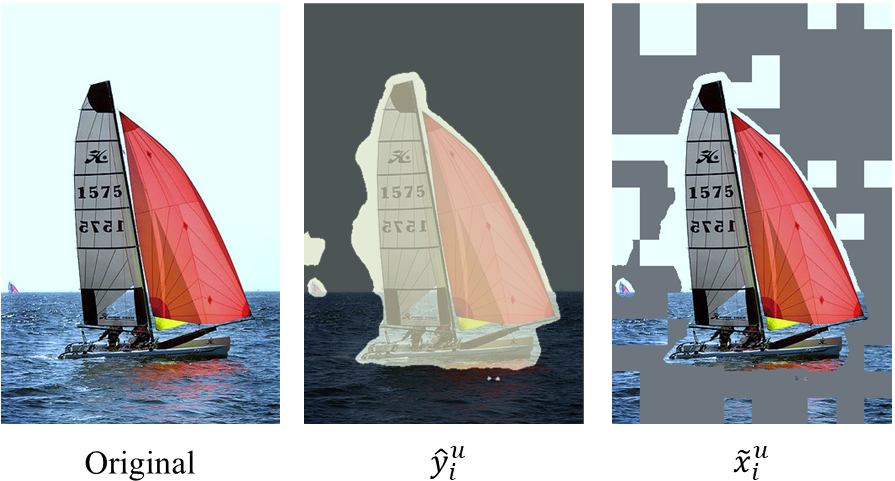}
   \caption{Illustration of Trusted Mask Perturbation. The white area in $\hat{y}_i^u$ represents the unreliable region.}
   \label{fig:3}
\end{figure}

Without loss of generality, assuming that $c=2$ indicates the subset of predictions with high maximum confidence and residual dispersion, we define the projection matrix $Z=\Phi S_{n,2}^\ast$, representing the projection of predictions onto the reliable component. Based on this, applying a Gaussian weighting function to construct smooth loss weights $\omega_n$,
\begin{equation}
    \omega_n = \prod_{c} \exp\left(\frac{\left(h_{n}(c) - \mu_c\right)^2}{-\alpha \sigma_c^2}\right),
    \label{eq:7} 
\end{equation}
where \(\mu_c\) is the mean of the high-reliability prediction subset in the projection matrix \(Z\) for component \(c\), and \(\sigma_c\) is its variance. \(\alpha = 8\) is the smoothing parameter. To ensure that predictions with high 
maximum confidence and residual dispersion are always retained, the weights of these predictions are set to 1. Therefore, the unsupervised training indicator mapping $M_i$ for \(x_i^u\) is defined as:
\begin{equation}
    M_i(n) = 
    \begin{cases} 
        1 & \text{if } (h_n(c) - \mu_c) > 0 , \forall c \in \{1,2\} \\
        \omega_n & \text{otherwise}
    \end{cases}.
    \label{eq:8}
\end{equation}

\begin{table*}[t]
  \centering
    \begin{tabularx}{\hsize}{l*{6}{>{\centering\arraybackslash}X}}
    \toprule[1.1pt]
    \rowcolor{color1}
    \textbf{PASCAL} \textit{original} & {Train Size} & 1/16(92) & 1/8(183) & 1/4(366) & 1/2(732)  & Full(1464) \\
    \midrule[0.1pt]
    Supervised& {321 × 321} & 45.4 & 53.1  & 61.2  & 68.4  & 73.2 \\
    ST++~\cite{ST++}      & {321 × 321} & 65.2 & 71.0  & 74.6  & 77.3  & 79.1 \\
    UniMatch~\cite{UniMatch}  & {321 × 321} & 75.2 & 77.2  & 78.8  & 79.9  & 81.2 \\
    CorrMatch~\cite{CorrMatch} & {321 × 321} & \underline{76.4} & \underline{78.5}  & 79.4  & 80.6  & 81.8 \\
    ESL~\cite{ESL}     & {513 × 513} & 71.0 & 74.1  & 78.1  & 79.5  & 81.8 \\
    LogicDiag~\cite{LOGICDIAG} & {513 × 513} & 73.3 & 76.7  & 78.0  & 79.4  & -    \\
    RankMatch~\cite{rankmatch} & {513 × 513} & 75.5 & 77.6  & \underline{79.8}  & 80.7 & \underline{82.2} \\
    AugSeg~\cite{augseg}  & {513 × 513} & 71.1 & 75.5  & 78.8  & 80.3  & 81.4 \\
    SemiCVT~\cite{Semicvt}   & {513 × 513} & 68.6 & 71.3  & 75.0  & 78.5  & 80.3 \\
    FPL~\cite{FPL}   & {513 × 513} & 69.3 & 71.7  & 75.7  & 79.0  & -    \\
    AllSpark~\cite{allspark} & {513 × 513} & 76.1 & 78.4  & 79.8  & \underline{80.8}  & 82.1 \\
    \midrule
    \rowcolor{color2}
    CSL       & {321 × 321} & \textbf{76.8} & \textbf{79.6} & \textbf{80.3} & \textbf{80.9} & \textbf{82.3} \\
    \bottomrule[1.1pt]
    \end{tabularx}
  \caption{Comparison with SOTAs on the PASCAL VOC 2012 original dataset under different splits. All labelled images were selected from the original VOC training set, which consists of 1,464 images. The fractions denote the percentage of labeled data used for training.}
  \label{tab:1}
\end{table*}

\subsection{Trusted Mask Perturbation}
\label{subsec:3.4}

While semi-supervised training can benefit from high-quality pseudo-labels, persistent exclusion of challenging regions risks entrenching the model in simplistic patterns. To mitigate this, we propose the Trusted Mask Perturbation (TMP), which selectively masks high-confidence predictions, thereby implicitly supervising unreliable regions using contextual information.

Utilizing the reliable pixel indicator \( M_i \) from \cref{subsec:3.3}, where \( M_i = 1 \) denotes reliable predictions, we define a patch-based perturbation mask \( R_i \in \{0, 1\}^{H \times W} \) generated as follows:
\begin{equation}
    R_i(x, y) = \mathbb{I}\bigl(M_i(x, y)=1\bigr) \cdot \mathbb{I}\bigl( U( \lfloor \tfrac{x}{s} \rfloor, \lfloor \tfrac{y}{s} \rfloor ) \geq \theta \bigr),
    \label{eq:9}
\end{equation}
where $s=32$ is the patch size, $\theta=0.7$ is the masking ratio, and $U(\cdot, \cdot)$ denotes the uniform distribution over the interval $[0, 1]$.

The perturbed image $\tilde{x}_i^u$ is obtained by applying $R_i$ to the input image $x_i^u$:
\begin{equation}
    \tilde{x}_i^u = x_i^u \odot R_i, 
    \label{eq:10}
\end{equation}
where $\odot$ denotes element-wise multiplication. As illustrated in \cref{fig:3}, this strategy selectively masks reliable pixels, compelling the model to infer the missing content based on contextual cues from the remaining challenging regions. 

The unsupervised loss for the masked inputs is formulated as:
\begin{equation}
    \mathcal{L}_u^m = \frac{1}{B_U} \sum_{i=1}^{B_U} M_i \odot CE\left( \hat{y}_i^u, F\left( \tilde{x}_i^u \right) \right). 
    \label{eq:11}
\end{equation}

The final unsupervised loss term combines both the standard and masked losses:
\begin{equation}
    \mathcal{L}_U =  \lambda_1L_u^a + \lambda_2L_u^m , 
    \label{eq:12}
\end{equation}
where $\lambda_1, \lambda_2$ are loss weights set to $[0.5, 0.5]$ by default.
\section{Experiments}
\label{sec:experiments}
\subsection{Implementation Details}
\label{subsec:4.1}
\begin{table*}[t]
  \centering
    \begin{tabularx}{\columnwidth}{p{2.3cm}c*{3}{>{\centering\arraybackslash}X}} 
      \specialrule{1.1pt}{0pt}{0pt}
      \rowcolor{color1}
      & & 1/16 & 1/8 & 1/4 \\
      \rowcolor{color1}
      \multirow{-2}{*}{\textbf{PASCAL} \textit{blender}} & \multirow{-2}{*}{Train Size} & (662) & (1323) & (2646) \\
      \specialrule{0.1pt}{0pt}{0.1pt}
      ST++~\cite{ST++}      & 321 × 321  & {74.5} & 76.3  & 76.6 \\
      UniMatch~\cite{UniMatch}   & 321 × 321  & {76.5} & 77.0  & 77.2 \\
      CorrMatch~\cite{CorrMatch}  & 321 × 321  & \underline{77.6} & \underline{77.8}  & \underline{78.3} \\
      CAC~\cite{CAC}        & 321 × 321  & {72.4} & 74.6  & 76.3 \\
      \midrule
      \rowcolor{color2}
      CSL        & 321 × 321 & {\textbf{77.8}} & \textbf{78.5} & \textbf{79.0} \\
      \midrule
      \midrule
      U2PL$^\dag$~\cite{U2PL}       & 513 × 513 & {77.2} & 79.0  & 79.3 \\
      GTA$^\dag$~\cite{GTA}      & 513 × 513 & {77.8} & 80.4  & 80.5 \\
      CorrMatch$^\dag$~\cite{CorrMatch}  & 513 × 513 & {81.3} & 81.9  & \underline{80.9} \\
      UniMatch$^\dag$~\cite{UniMatch}  & 513 × 513 & {81.0} & 81.9  & 80.4 \\
      AugSeg$^\dag$~\cite{augseg}   & 513 × 513 & {79.3} & 81.5  & 80.5 \\
      AllSpark$^\dag$~\cite{allspark}  & 513 × 513 & \underline{81.6} & \underline{82.0}  & 80.9 \\
      \midrule
      \rowcolor{color2}
      CSL$^\dag$       & 513 × 513 & \textbf{81.6} & \textbf{82.4} & \textbf{81.1} \\
      \bottomrule[1.1pt]
    \end{tabularx}
  \hfill
    \begin{tabularx}{\columnwidth}{p{2.3cm}c*{3}{>{\centering\arraybackslash}X}}
      \specialrule{1.1pt}{0pt}{0pt}
      \rowcolor{color1}
      & & 1/16 & 1/8 & 1/4 \\
      \rowcolor{color1}
      \multirow{-2}{*}{\textbf{PASCAL} \textit{blender}} & \multirow{-2}{*}{Train Size} & (662) & (1323) & (2646) \\
      \specialrule{0.1pt}{0pt}{0.1pt}
      ST++~\cite{ST++}     & 513 × 513 & {74.7} & 77.9  & 77.9 \\
      UniMatch~\cite{UniMatch} & 513 × 513 & {78.1} & 78.4  & 79.2 \\
      CorrMatch~\cite{CorrMatch} & 513 × 513 & {78.4} & 79.3  & 79.6 \\
      ESL~\cite{ESL}      & 513 × 513 & {76.4} & 78.6  & 79.0 \\
      PS-MT~\cite{PS-MT}       & 513 × 513 & {75.5} & 78.2  & 78.7 \\
      CFCG~\cite{CFCG}    & 513 × 513 & {76.8} & 79.1  & 80.0 \\
      CCVC~\cite{CCVC}      & 513 × 513 & {77.2} & 78.4  & 79.0 \\
      DLG~\cite{DLG}       & 513 × 513 & {77.8} & 79.3  & 79.1 \\
      RankMatch~\cite{rankmatch}  & 513 × 513 & \underline{78.9} & 79.2    & 80.0 \\
      DGCL~\cite{DGCL}      & 513 × 513 & {76.6} & 78.3  & 79.3 \\
      AllSpark~\cite{allspark} & 513 × 513 & {78.3} & \textbf{80.0}  & \textbf{80.4} \\
      DDFP~\cite{DDFP}  & 513 × 513 & {78.3} & 78.9    & 79.8 \\
      \midrule
      \rowcolor{color2}
      CSL       & 513 × 513 & \textbf{78.9} & \underline{79.9} & \underline{80.3} \\
      \bottomrule[1.1pt]
    \end{tabularx}
  \caption{Comparison with SOTAs on PASCAL VOC 2012 blender dataset. All labelled images were selected from the blender VOC training set under different splits, which consists of 10,582 images. $^\dag$ means using U2PL's splits.}
  \label{tab:2}
\end{table*}

\begin{table}[t]
  \centering
    \begin{tabularx}{\columnwidth}{p{1.6cm}*{4}{>{\centering\arraybackslash}X}}
    \toprule[1.1pt]
    \rowcolor{color1}
    \textbf{Cityscapes} & 1/16(186) & 1/8(732) & 1/4(744) & 1/2(1488)\\
    \midrule[0.1pt]
    Supervised & {63.1} & {70.5} & {73.1} & {76.2} \\
    ST++~\cite{ST++}    & {67.6} & {73.4} & {74.6} & {77.8} \\
    UniMatch~\cite{UniMatch}  & {76.6} & {77.9} & {79.2} & {79.5} \\
    CorrMatch~\cite{CorrMatch}  & \underline{77.3} & {78.5} & {79.4} & {80.4} \\
    ESL ~\cite{ESL}     & {75.1} & {77.2} & {79.0} & {80.5} \\
    DGCL~\cite{DGCL}    & {73.2} & {77.3} & {78.5} & {80.7} \\
    AugSeg~\cite{augseg}   & {75.2} & {77.8} & {79.6} & {80.4} \\
    CCVC~\cite{CCVC}    & {74.9} & {76.4} & {77.3} & {-} \\
    DAW~\cite{DAW}     & {76.6} & {78.4} & {79.8} & {80.6} \\
    U2PL~\cite{U2PL}      & {70.3} & {74.4} & {76.5} & {79.1} \\
    CPS~\cite{cps}    & {69.8} & {74.3} & {74.6} & {76.8} \\
    DDFP~\cite{DDFP}  & 77.1 & 78.2 & \underline{79.9} & \underline{80.8} \\
    FPL~\cite{FPL}    & {75.7} & \underline{78.5} & {79.2} & {-} \\
    \midrule
    \rowcolor{color2}
    CSL   & \textbf{78.2} & {\textbf{78.8}} & {\textbf{80.0}} & {\textbf{81.1}} \\
    \bottomrule[1.1pt]
    \end{tabularx}
  \caption{Comparison with SOTAs on Cityscapes dataset. All methods are built upon DeepLabV3+~\cite{deeplabv3+} and ResNet101~\cite{resnet}.}
  \label{tab:3}
\end{table}

\noindent\textbf{Datasets.} We evaluate the proposed method on three benchmark datasets: PASCAL VOC 2012, Cityscapes, and MS COCO. PASCAL VOC 2012~\cite{pascal} is a standard semantic segmentation dataset comprising 20 object classes and one background class. The training and validation sets contain 1,464 and 1,449 images, respectively. Following previous works~\cite{pseudoseg,GTA}, we augment the training set by incorporating 9,118 images with coarse annotations from the Segmentation Boundary Dataset (SBD)~\cite{SBD}, resulting in a blender set of 10,582 training images. Cityscapes~\cite{cityscapes} focuses on urban scene understanding and includes 19 classes. It consists of 2,975 finely annotated training images and 500 validation images, all at a high resolution of 1024×2048. MS COCO~\cite{coco} is a large-scale dataset with 81 object categories, containing 118k training and 5k validation images for segmentation tasks.

\noindent\textbf{Implementation Details.} To ensure a fair comparison with prior methods, we adopt the ResNet~\cite{resnet} backbone with DeepLabv3+~\cite{deeplabv3+} as the decoder for PASCAL VOC 2012 and Cityscapes, and Xception-65~\cite{xception} for MS COCO~\cite{coco}. For all datasets, we use SGD with a batch size of 8 and weight decay of $1\times10^{-4}$. For PASCAL VOC 2012, we set the initial learning rate to 0.001, crop size of 321×321 or 513×513, training epochs of 80, and a decoder learning rate ten times that of the backbone. For Cityscapes, we use an initial learning rate of 0.001, crop size of 801×801, and training epochs of 240, with OHEM~\cite{OHEM}. For COCO, we set the initial learning rate to 0.004, crop size of 513×513 and train for 30 epochs.

\noindent\textbf{Evaluation Metrics.} For PASCAL VOC 2012, we follow standard protocols~\cite{DARS,U2PL}, reporting mIoU on the original validation images. For Cityscapes, we perform sliding window evaluation with a crop size of 801×801 and compute mIoU~\cite{DLG,AEL}. For COCO, we evaluate on the standard 5k validation set with mIoU. Ablation studies are conducted on PASCAL VOC 2012 with a crop size of 321×321.

\subsection{Comparison with State-of-the-Art Methods}

\noindent\textbf{Original PASCAL VOC 2012.} \cref{tab:1} presents the comparison results on the original PASCAL VOC 2012 dataset. Compared to the supervised baseline, CSL consistently achieves significant performance improvements. Specifically, under the 1/16, 1/8, 1/4, 1/2, and full data splits, CSL attains improvements of +31.4\%, +26.5\%, +19.1\%, +12.5\%, and +9.1\%, respectively. Moreover, when compared with existing SOTAs, our approach consistently demonstrates superior performance. Particularly, under the 1/8 and 1/4 splits, CSL outperforms UniMatch~\cite{UniMatch} by +2.4\% and +1.5\%, respectively.

\begin{figure*}[t]
  \centering
  \begin{subfigure}[b]{0.33\linewidth}
    \includegraphics[width=\linewidth]{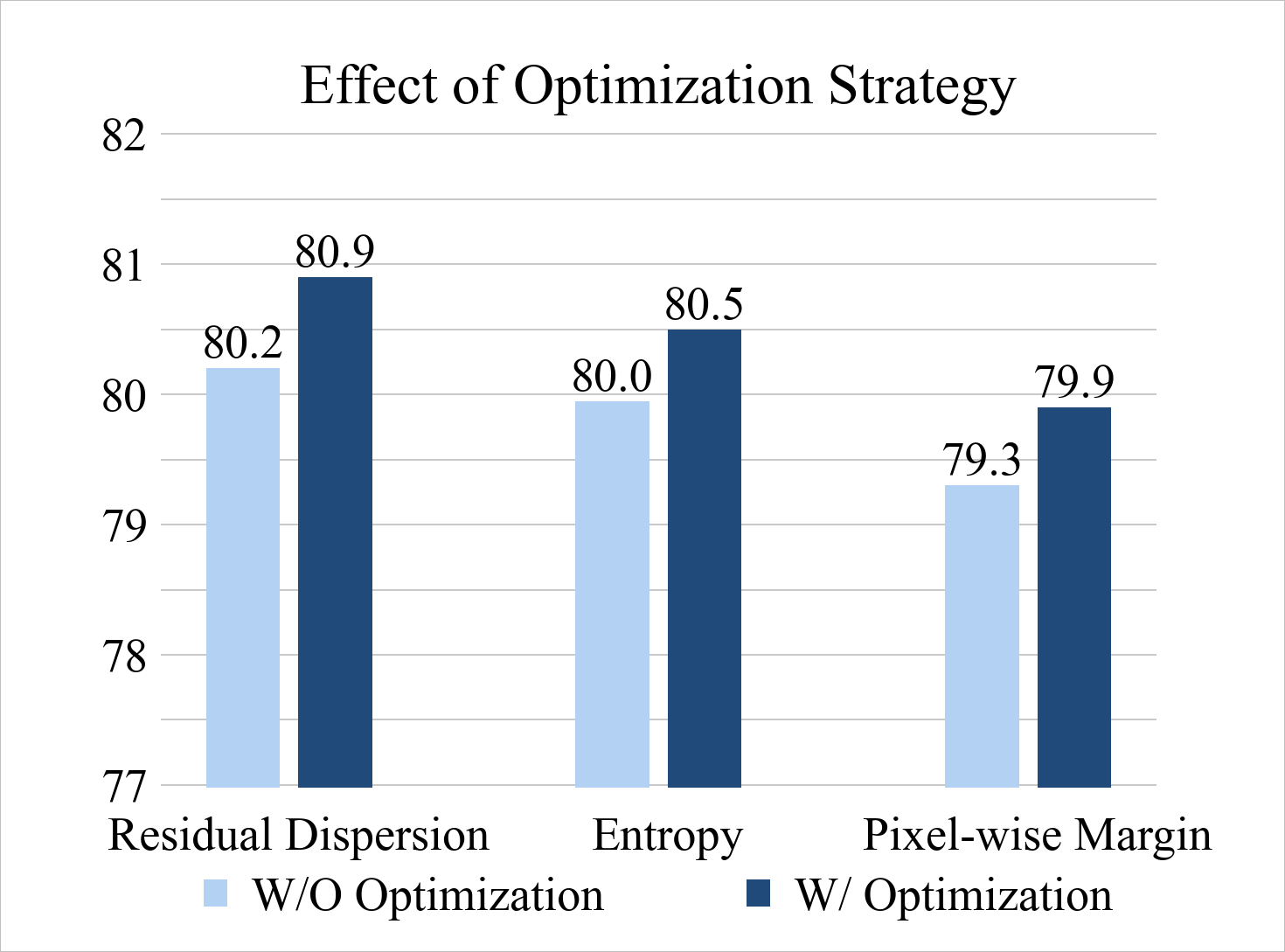}
    \caption{Dispersion Metrics}
    \label{fig:4a}
  \end{subfigure}
  \begin{subfigure}[b]{0.33\linewidth}
    \includegraphics[width=\linewidth]{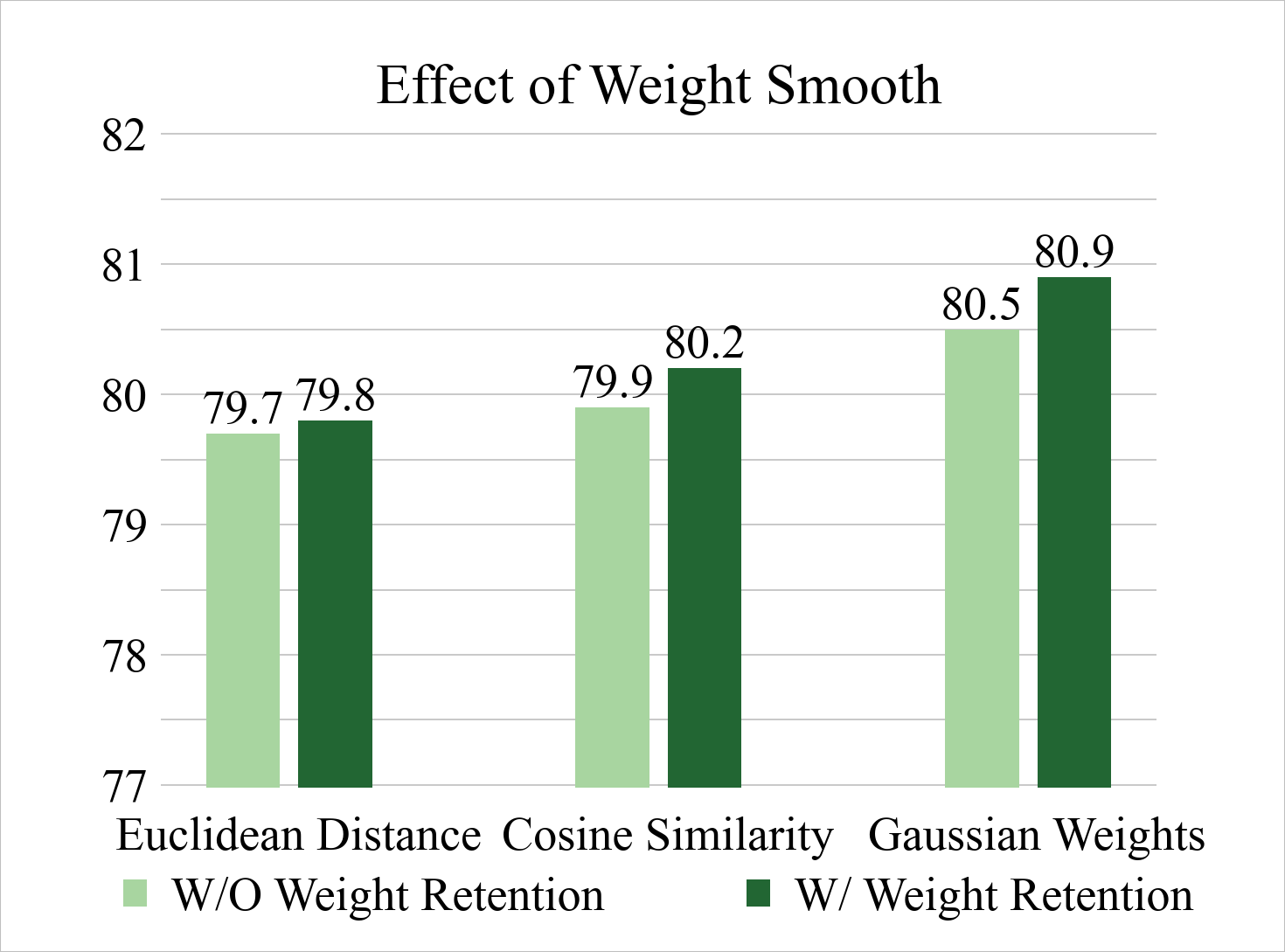}
    \caption{Loss Smoothing Strategy}
    \label{fig:4b}
  \end{subfigure}
  \begin{subfigure}[b]{0.33\linewidth}
    \includegraphics[width=\linewidth]{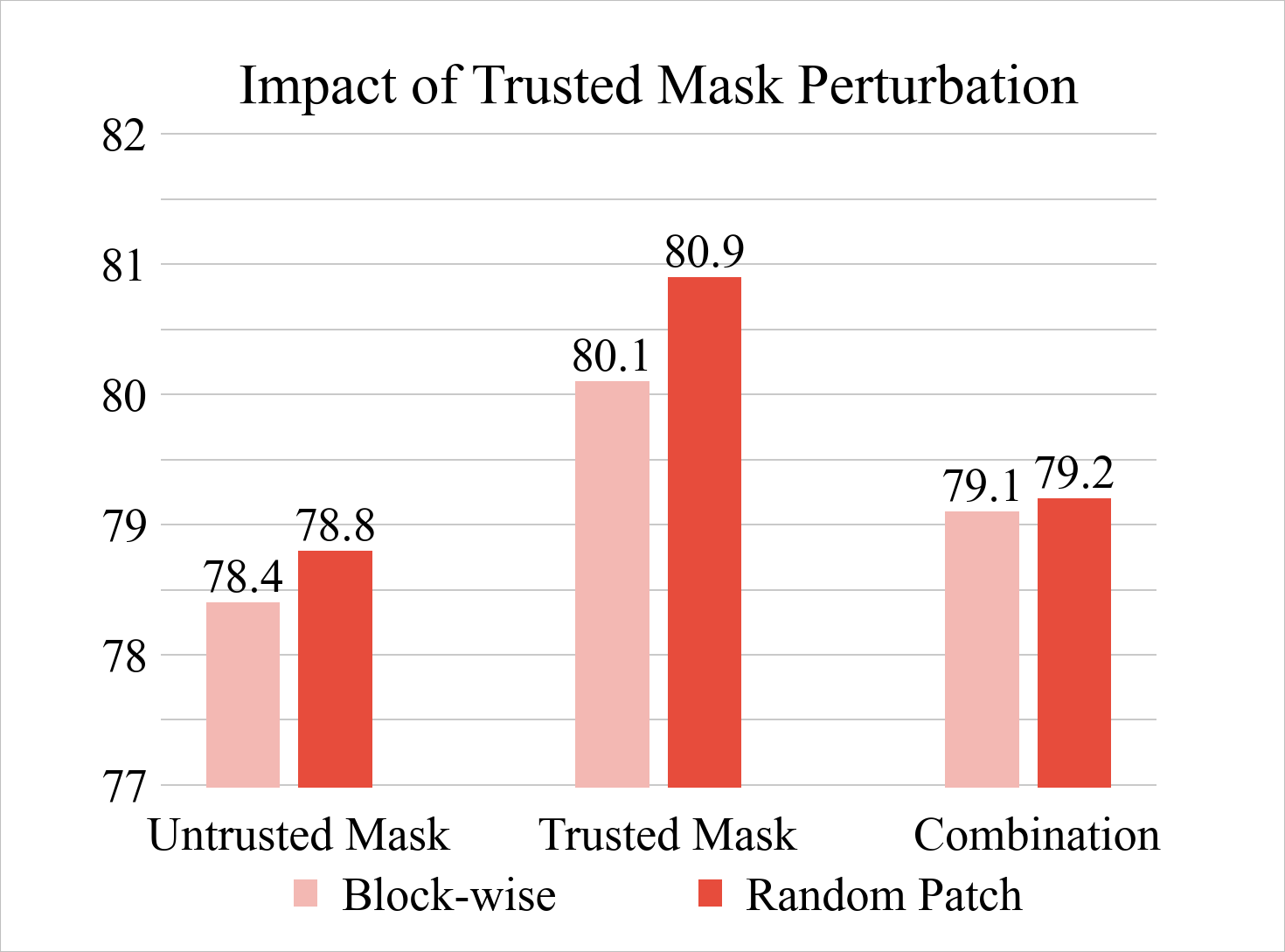}
    \caption{Trusted Mask Perturbation}
    \label{fig:4c}
  \end{subfigure}
  \caption{Ablation studies on different designs of CSL. For (a), (b), and (c), experiments are conducted on the 1/2 splits.}
  \label{fig:4}
\end{figure*}

\noindent\textbf{Blender PASCAL VOC 2012.} We show the comparison between our method and SOTAs on the blender PASCAL VOC 2012 dataset in \cref{tab:2}. At a crop size of 321×321, CSL surpasses CorrMatch~\cite{CorrMatch} by +0.2\%, +0.7\%, and +0.7\%. We observe that the performance gains become more pronounced as the amount of available annotations decreases. For instance, at a crop size of 513×513 under the 1/16 splits, our method improves upon AllSpark~\cite{allspark} by +0.6\%.

Additionally, we conduct experiments using the same splits as U2PL~\cite{U2PL}, where the supervised dataset primarily consists of high-quality annotations supplemented with a small amount of coarse annotation data. Under the 1/8 splits, since most high-quality annotations are included and the coarse annotations are largely excluded, the model benefits from a robust initial representation. This enables our method to achieve an outstanding performance of 82.4\%, representing a +0.9\% improvement over AugSeg~\cite{augseg}. This significant performance gain demonstrates our method's ability to effectively leverage accurate network representations in scenarios with scarce annotations.

\begin{table}[t]
  \small
  \centering
  \setlength{\tabcolsep}{2.4pt}
    \begin{tabularx}{\columnwidth}{p{1.6cm}*{5}{>{\centering\arraybackslash}X}}
    \toprule[1.1pt]
    \rowcolor{color1}
    \textbf{COCO} & 1/512 & 1/256 & 1/128 & 1/64 & 1/32\\
    \midrule[0.1pt]
    Supervised & {22.5} & {28.3} & {33.1} & {38.9} & {42.1} \\
    UniMatch~\cite{UniMatch}   & {31.9} & {38.9} & {44.4} & {48.2} & {49.8}  \\
    LogicDiag~\cite{LOGICDIAG}  & \underline{33.1} & \underline{40.3} & \underline{45.4} & \underline{48.8} & \underline{50.5}  \\
    PseudoSeg~\cite{pseudoseg}  & {29.8} & {37.1} & {39.1} & {41.8} & {43.6}  \\
    \midrule
    \rowcolor{color2}
    CSL & \textbf{35.1} & \textbf{42.3} & \textbf{45.8} & \textbf{49.7} & \textbf{51.2} \\
    \bottomrule[1.1pt]
    \end{tabularx}
  \caption{Comparison with SOTAs on MS-COCO dataset. All methods are built upon XC-65~\cite{xception} for fair comparison.}
  \label{tab:coco}
\end{table}

\noindent\textbf{Cityscapes.} \cref{tab:3} quantitatively compares the performance of CSL with SOTAs on the Cityscapes validation set. Leveraging the extensive unlabeled data, CSL achieves performance gains of +15.1\%, +8.3\%, +6.9\%, and +5.1\% over the supervised baseline under the 1/16, 1/8, 1/4, and 1/2 splits, respectively. CSL surpasses existing SOTAs across all partition protocols. For example, under the 1/16 partition, our method achieves a +0.9\% performance improvement over CorrMatch~\cite{CorrMatch}.

\noindent\textbf{MS COCO.} \cref{tab:coco} presents the quantitative comparisons of CSL with SOTAs on the MS COCO dataset. Across all partition ratios, CSL consistently outperforms existing approaches by significant margins. Notably, when utilizing only 1/512 of the labeled data, CSL achieves an impressive 35.1\% mIoU, surpassing the strongest baseline LogicDiag~\cite{LOGICDIAG} by +2.0\%. 

\subsection{Ablation Studies}

\begin{table}[t]
  \centering
  \begin{tabularx}{\columnwidth}{c|*{3}{>{\centering\arraybackslash}X}|*{1}{>{\centering\arraybackslash}X}|*{1}{>{\centering\arraybackslash}X}}
    \toprule[1.1pt]
    \multirow{2}{*}{SemiBaseline} & \multicolumn{3}{c|}{CSL} & 1/16 & 1/2 \\
    & $M_i$ & $\omega_n$ & $\mathcal{L}_u^m$ & (92) & (732) \\
    \midrule
    \checkmark &       &       &       & 72.9 & 78.6 \\
    \checkmark & \checkmark     &       &       & 73.8 & 79.3 \\
    \checkmark & \checkmark     & \checkmark     &       & 74.2 & 79.8 \\
    \checkmark &       &       & \checkmark     & 73.6 & 79.4 \\
    \checkmark & \checkmark     & \checkmark     & \checkmark     & \textbf{76.8} & \textbf{80.9} \\
    \bottomrule[1.1pt]
  \end{tabularx}
    \caption{Ablation study on the effectiveness of different components, including optimization strategy $M_i$, smooth loss weight $\omega_n$, trusted mask perturbation $\mathcal{L}_u^m$.}
  \label{tab:4}
\end{table}

\noindent\textbf{Effectiveness of components.} We conducted ablation studies on the different components of CSL to demonstrate their effectiveness, as shown in \cref{tab:4}. Starting with a semi-supervised baseline, we achieve mIoU scores of 72.9\% and 78.6\% for the 1/16 and 1/2 splits, respectively. By introducing the residual dispersion and optimization strategy $M_i$, the performance improves to 73.8\% and 79.3\%. Building upon this, incorporating the smooth loss weight $\omega_n$ yields additional gains of 0.4\% and 0.5\%. By combining all components, our method surpasses the baseline by over 3.9\% and 2.3\%, achieving SOTA results. Notably, adding the trusted mask perturbation $\mathcal{L}_u^m$ alone to the semi-supervised baseline results in performance gains of 0.7\% and 0.8\%, illustrating its general applicability in addressing the lack of supervision in challenging regions of pseudo-labels.

\noindent\textbf{Evaluation of Dispersion Metrics.} \cref{fig:4a} compares residual dispersion with pixel-wise margin~\cite{shin2021all} and entropy. Both margin and entropy exhibit significant performance degradation. Because when the model is overconfident, the margin dominates by the maximum confidence, while entropy becomes insensitive to non-maximum confidence values. Under such circumstances, reliability assessment degenerates into merely comparing maximum confidence scores. Following the setup in \cref{subsec:3.2}, residual dispersion maintains magnitude consistency in scale and coefficients, ensuring robust pseudo-label selection. Meanwhile, replacing the optimization strategy with sample-wise mean thresholds~\cite{softmatch} resulted in consistent performance degradation, validating the necessity of our design.

\begin{figure}[t]
  \centering
   \includegraphics[width=0.85\linewidth]{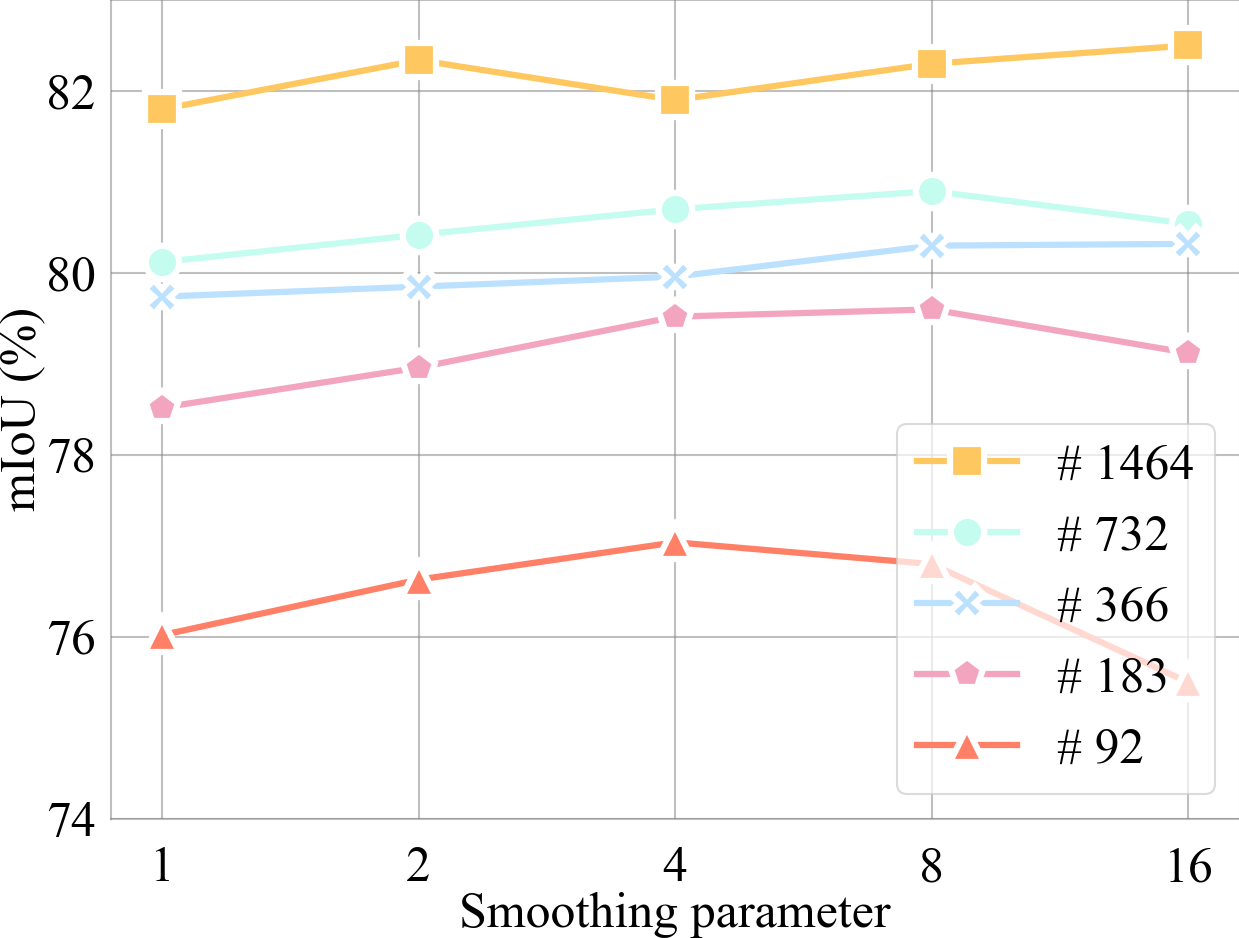}
   \caption{Ablation study of smoothing parameter $\alpha$.}
   \label{fig:5}
\end{figure}

\noindent\textbf{Weight smoothing strategy.} CSL constructs smoothed loss weights by the Gaussian weighting function. We compare it with alternative measures, including Euclidean distance and cosine similarity. Additionally, we examined the effect of retaining loss weights for reliable predictions versus uniformly smoothing all loss weights. As shown in \cref{fig:4b}, combining Gaussian weighting with retention of reliable prediction loss weights yields the most significant performance improvement. Suggests that our approach ensures models learn effectively from credible examples, enhancing overall segmentation performance.

\noindent\textbf{Impact of trusted mask perturbation.} \cref{fig:4c} analyses the performance of block-wise masking~\cite{baobeit} versus random patch masking across different masking targets. The results show that masking only reliable predictions yields the best performance. In contrast, masking untrustworthy predictions or both results in significant performance degradation. This decline may stem from the absence of supervised signals and image content in untrustworthy regions, leading to excessive perturbation. Furthermore, random patch masking consistently outperforms block-wise masking due to the increased sample diversity introduced.

\noindent\textbf{Impact of hyperparameters.} The effects of the loss smoothing coefficient $\alpha$, masking patch size $s$, and masking rate $\theta$ on the PASCAL VOC 2012 dataset were investigated. \cref{fig:5} illustrates that an excessively high value of $\alpha$ leads to performance degradation, which results from the improper weighting of unreliable predictions. Consequently, $\alpha$ = 8 has been established as the default value to optimize overall performance. \cref{tab:6} demonstrates that optimal performance enhancement is achieved from credible masking when $\theta$ = 0.7 and $s$ = 32. In contrast, when the masking patch size $s$ = 64, the mIoU decreases significantly, which is attributed to the extreme size of the patch compared to the image size, ultimately leading to a loss of local contextual information.

\subsection{Qualitative analysis}

\begin{figure}[t]
  \centering
  \begin{subfigure}[b]{0.495\linewidth}
    \includegraphics[width=\linewidth]{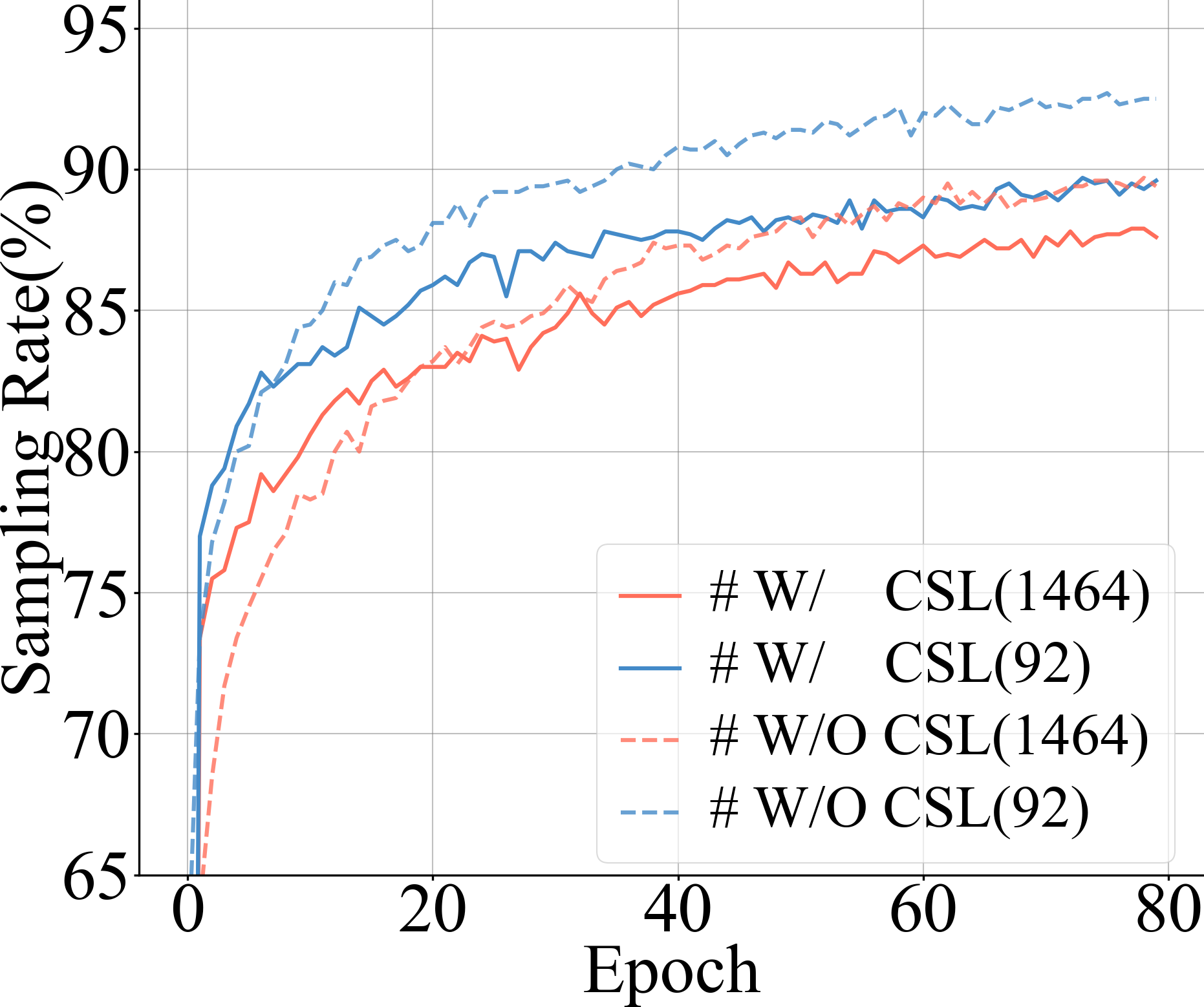}
    \caption{Sampling Rate}
    \label{fig:6a}
  \end{subfigure}
  \begin{subfigure}[b]{0.495\linewidth}
    \includegraphics[width=\linewidth]{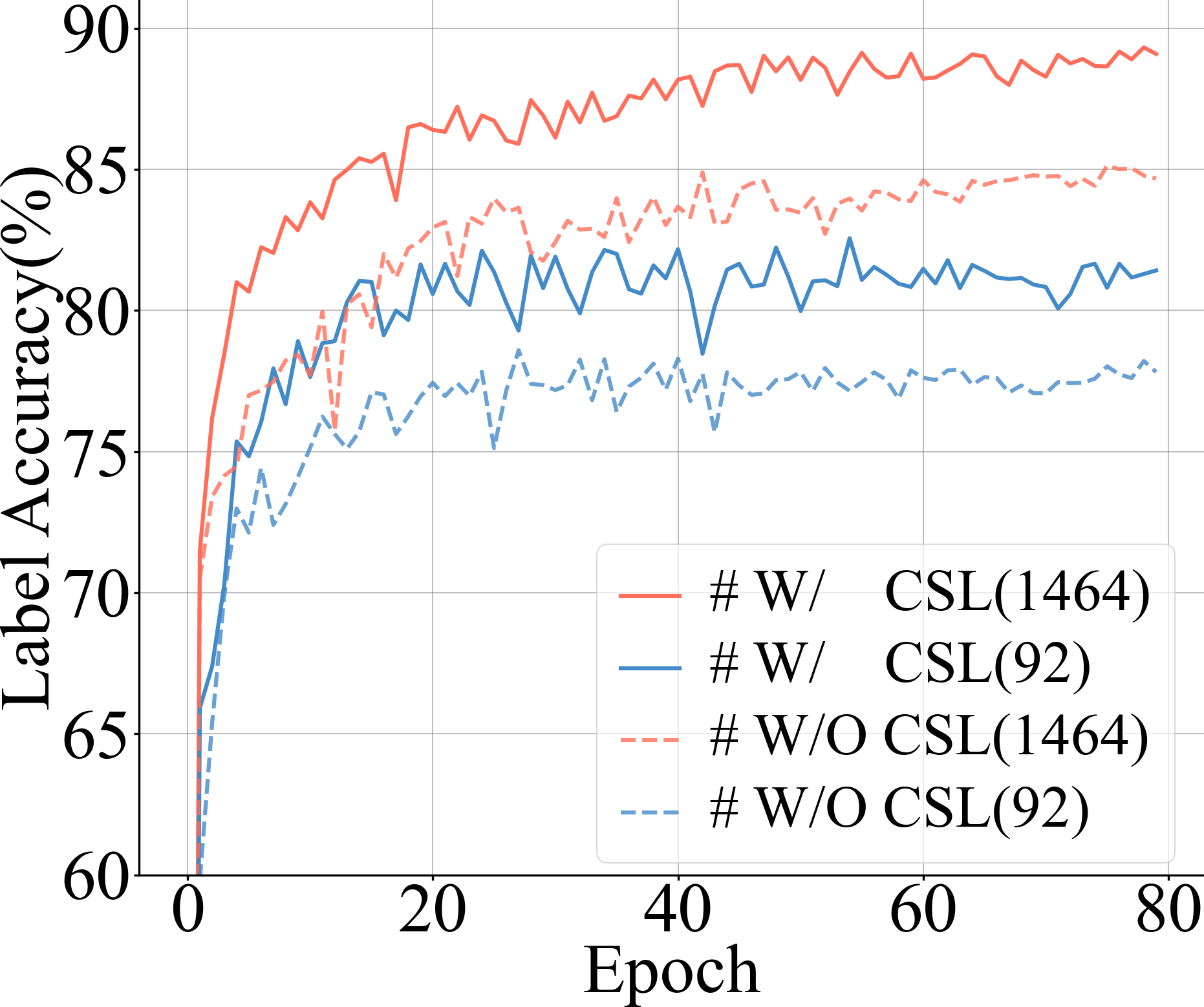}
    \caption{Pseudo-label accuracy}
    \label{fig:6b}
  \end{subfigure}
  \caption{Training curves for our CSL and the threshold strategy.}
  \label{fig:6}
\end{figure}

\begin{table}[t]
  \centering
  \begin{tabularx}{\columnwidth}{c|*{4}{>{\centering\arraybackslash}X}}
    \toprule[1.1pt]
    Patch Size $s$ & 8 & 16 & 32 & 64\\
    \cmidrule{1-5}
    $\theta$=0.3 & 78.7 & 78.5 & 78.4          & 76.2\\
    $\theta$=0.5 & 78.2 & 79.1 & 78.9          & 77.8\\
    $\theta$=0.7 & 78.8 & 79.2 & \textbf{79.4} & 78.4\\
    $\theta$=0.9 & 77.5 & 77.6 & 77.6          & 77.4\\
    \bottomrule[1.1pt]
  \end{tabularx}
  \caption{Ablation study with different mask perturbation parameters for 1/2 splits.}
  \label{tab:6}
\end{table}

\noindent\textbf{CSL improved semi-supervised training.} To evaluate the effectiveness of CSL, we measured the average accuracy and the average sampling rate of pseudo-labels produced by our method and threshold-based methods. As seen in \cref{fig:6}, suffering from overconfidence, as the quantity of labeled data diminishes, threshold-based methods exhibit a proclivity for increasingly assertive pseudo-label selection, resulting in a notable surge in false supervision signals. In contrast, CSL demonstrates superior performance in mitigating cognitive biases and consistently generates high-quality pseudo-labels.

\begin{figure}[t]
  \centering
  \begin{subfigure}[b]{0.30\linewidth}
    \includegraphics[width=\linewidth]{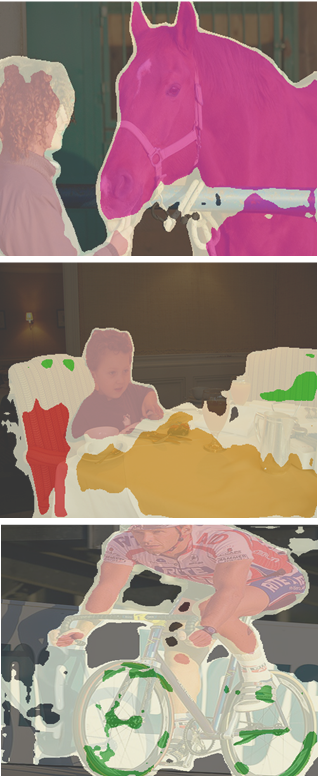}
    \caption{W/O CSL}
    \label{fig:7a}
  \end{subfigure}
  \begin{subfigure}[b]{0.30\linewidth}
    \includegraphics[width=\linewidth]{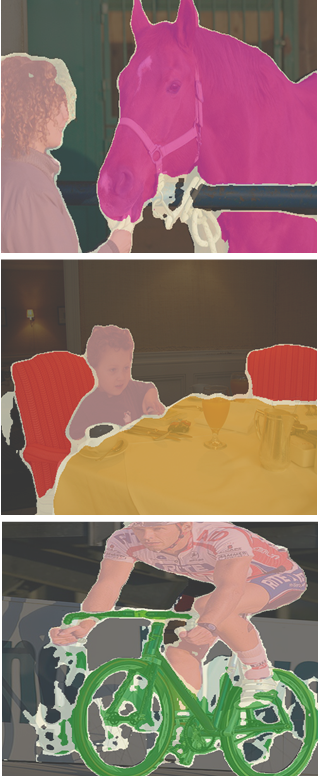}
    \caption{W/ CSL}
    \label{fig:7b}
  \end{subfigure}
  \begin{subfigure}[b]{0.30\linewidth}
    \includegraphics[width=\linewidth]{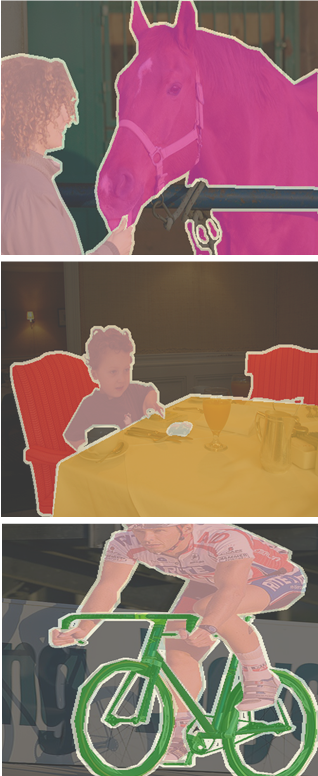}
    \caption{GT}
    \label{fig:7c}
  \end{subfigure}
  \caption{Qualitative results on the PASCAL VOC 2012 dataset. (a) Pseudo labels without CSL; (b) Pseudo labels with CSL; (c) Ground truth.}
  \label{fig:7}
\end{figure}

\noindent\textbf{Qualitative Results.} \cref{fig:7} illustrates the visualization results for the PASCAL VOC 2012 dataset under 1/8 splits.  Obviously, our predictive convex optimized segmentation strategy effectively broadens the regions encompassed by pseudo-labels. By utilizing plausible mask perturbations, CSL achieves more accurate and detailed segmentation in intricate areas, and the visualization results demonstrate its capacity to enhance segmentation performance by leveraging unlabelled data and addressing supervision gaps.
\section{Conclusions}
\label{sec:conclusions}

We demonstrate that through comprehensive measurement of maximum confidence and residual dispersion, a pronounced natural separation exists between reliable predictions and cognitive biases. Motivated by this, we explore pseudo-label selection in this space through a convex optimization strategy. On the other hand, we attempt to motivate the supervised signals of reliable predictions to improve prediction results in unreliable regions by masking the reliable pixels. Experiments demonstrate the significant effectiveness of decision boundaries in this space, and that masking strategy significantly improves prediction results in complex regions. In addition, comparative experiments demonstrate the superiority of CSL over other methods.

\noindent\textbf{Acknowledgements.} This work is supported by the China Postdoctoral Program for State-Owned Enterprises (GZC20251179) and Hunan Youth Science Foundation Program (2025JJ60424).

{
    \small
    \bibliographystyle{ieeenat_fullname}
    \bibliography{main}

\begin{thebibliography}{76}
\providecommand{\natexlab}[1]{#1}
\providecommand{\url}[1]{\texttt{#1}}
\expandafter\ifx\csname urlstyle\endcsname\relax
  \providecommand{\doi}[1]{doi: #1}\else
  \providecommand{\doi}{doi: \begingroup \urlstyle{rm}\Url}\fi

\bibitem[Arazo et~al.(2020)Arazo, Ortego, Albert, O’Connor, and McGuinness]{bias}
Eric Arazo, Diego Ortego, Paul Albert, Noel~E O’Connor, and Kevin McGuinness.
\newblock Pseudo-labeling and confirmation bias in deep semi-supervised learning.
\newblock In \emph{IJCNN}, pages 1--8. IEEE, 2020.

\bibitem[Bao et~al.(2021)Bao, Dong, Piao, and Wei]{baobeit}
Hangbo Bao, Li Dong, Songhao Piao, and Furu Wei.
\newblock Beit: Bert pre-training of image transformers.
\newblock In \emph{ICLR}, 2021.

\bibitem[Chapelle et~al.(2009)Chapelle, Scholkopf, and Zien]{survey}
Olivier Chapelle, Bernhard Scholkopf, and Alexander Zien.
\newblock Semi-supervised learning.
\newblock \emph{IEEE Transactions on Neural Networks}, 20\penalty0 (3):\penalty0 542--542, 2009.

\bibitem[Chen et~al.(2022)Chen, Jiang, Wang, Wan, Wang, and Long]{chen2022debiased}
Baixu Chen, Junguang Jiang, Ximei Wang, Pengfei Wan, Jianmin Wang, and Mingsheng Long.
\newblock Debiased self-training for semi-supervised learning.
\newblock \emph{NeurIPS}, 35:\penalty0 32424--32437, 2022.

\bibitem[Chen et~al.(2023)Chen, Tao, Fan, Wang, Wang, Schiele, Xie, Raj, and Savvides]{softmatch}
Hao Chen, Ran Tao, Yue Fan, Yidong Wang, Jindong Wang, Bernt Schiele, Xing Xie, Bhiksha Raj, and Marios Savvides.
\newblock Softmatch: Addressing the quantity-quality tradeoff in semi-supervised learning.
\newblock In \emph{ICLR}, 2023.

\bibitem[Chen et~al.(2017)Chen, Papandreou, Kokkinos, Murphy, and Yuille]{deeplab}
Liang-Chieh Chen, George Papandreou, Iasonas Kokkinos, Kevin Murphy, and Alan~L Yuille.
\newblock Deeplab: Semantic image segmentation with deep convolutional nets, atrous convolution, and fully connected crfs.
\newblock \emph{IEEE TPAMI}, 40\penalty0 (4):\penalty0 834--848, 2017.

\bibitem[Chen et~al.(2018)Chen, Zhu, Papandreou, Schroff, and Adam]{deeplabv3+}
Liang-Chieh Chen, Yukun Zhu, George Papandreou, Florian Schroff, and Hartwig Adam.
\newblock Encoder-decoder with atrous separable convolution for semantic image segmentation.
\newblock In \emph{ECCV}, pages 801--818, 2018.

\bibitem[Chen et~al.(2021)Chen, Yuan, Zeng, and Wang]{cps}
Xiaokang Chen, Yuhui Yuan, Gang Zeng, and Jingdong Wang.
\newblock Semi-supervised semantic segmentation with cross pseudo supervision.
\newblock In \emph{CVPR}, pages 2613--2622, 2021.

\bibitem[Chen et~al.(2025)Chen, Shi, Zhou, He, and Tsoka]{Seg1}
Xinyue Chen, Miaojing Shi, Zijian Zhou, Lianghua He, and Sophia Tsoka.
\newblock Enhancing generalized few-shot semantic segmentation via effective knowledge transfer.
\newblock In \emph{AAAI}, 2025.

\bibitem[Cheng et~al.(2025)Cheng, Huang, Wang, Huang, and Wei]{Seg4}
Shengyang Cheng, Jianyong Huang, Xiaodong Wang, Lei Huang, and Zhiqiang Wei.
\newblock Image-text aggregation for open-vocabulary semantic segmentation.
\newblock \emph{Neurocomputing}, page 129702, 2025.

\bibitem[Chollet(2017)]{xception}
Fran{\c{c}}ois Chollet.
\newblock Xception: Deep learning with depthwise separable convolutions.
\newblock In \emph{CVPR}, pages 1251--1258, 2017.

\bibitem[Cordts et~al.(2016)Cordts, Omran, Ramos, Rehfeld, Enzweiler, Benenson, Franke, Roth, and Schiele]{cityscapes}
Marius Cordts, Mohamed Omran, Sebastian Ramos, Timo Rehfeld, Markus Enzweiler, Rodrigo Benenson, Uwe Franke, Stefan Roth, and Bernt Schiele.
\newblock The cityscapes dataset for semantic urban scene understanding.
\newblock In \emph{CVPR}, pages 3213--3223, 2016.

\bibitem[Deng et~al.(2025)Deng, Lu, and Zhang]{Seg2}
Jiacheng Deng, Jiahao Lu, and Tianzhu Zhang.
\newblock Quantity-quality enhanced self-training network for weakly supervised point cloud semantic segmentation.
\newblock \emph{IEEE TPAMI}, 2025.

\bibitem[Du et~al.(2022)Du, Shen, Wang, Fei, Li, Wu, Zhao, Fu, and Liu]{FST}
Ye Du, Yujun Shen, Haochen Wang, Jingjing Fei, Wei Li, Liwei Wu, Rui Zhao, Zehua Fu, and Qingjie Liu.
\newblock Learning from future: A novel self-training framework for semantic segmentation.
\newblock \emph{NeurIPS}, 35:\penalty0 4749--4761, 2022.

\bibitem[Everingham et~al.(2015)Everingham, Eslami, Van~Gool, Williams, Winn, and Zisserman]{pascal}
Mark Everingham, SM~Ali Eslami, Luc Van~Gool, Christopher~KI Williams, John Winn, and Andrew Zisserman.
\newblock The pascal visual object classes challenge: A retrospective.
\newblock \emph{IJCV}, 111:\penalty0 98--136, 2015.

\bibitem[French et~al.(2020)French, Laine, Aila, and Mackiewicz]{cowmix}
G. French, S. Laine, T. Aila, and M. Mackiewicz.
\newblock Semi-supervised semantic segmentation needs strong, varied perturbations.
\newblock In \emph{BMVC}, 2020.

\bibitem[Guo et~al.(2017)Guo, Pleiss, Sun, and Weinberger]{calibration}
Chuan Guo, Geoff Pleiss, Yu Sun, and Kilian~Q Weinberger.
\newblock On calibration of modern neural networks.
\newblock In \emph{ICML}, pages 1321--1330. PMLR, 2017.

\bibitem[Hariharan et~al.(2011)Hariharan, Arbeláez, Bourdev, Maji, and Malik]{SBD}
Bharath Hariharan, Pablo Arbeláez, Lubomir Bourdev, Subhransu Maji, and Jitendra Malik.
\newblock Semantic contours from inverse detectors.
\newblock In \emph{ICCV}, pages 991--998, 2011.

\bibitem[He et~al.(2016)He, Zhang, Ren, and Sun]{resnet}
Kaiming He, Xiangyu Zhang, Shaoqing Ren, and Jian Sun.
\newblock Deep residual learning for image recognition.
\newblock In \emph{CVPR}, pages 770--778, 2016.

\bibitem[He et~al.(2021)He, Yang, and Qi]{DARS}
Ruifei He, Jihan Yang, and Xiaojuan Qi.
\newblock Re-distributing biased pseudo labels for semi-supervised semantic segmentation: A baseline investigation.
\newblock In \emph{CVPR}, pages 6930--6940, 2021.

\bibitem[Hong et~al.(2015)Hong, Noh, and Han]{Decoupled}
Seunghoon Hong, Hyeonwoo Noh, and Bohyung Han.
\newblock Decoupled deep neural network for semi-supervised semantic segmentation.
\newblock \emph{NeurIPS}, 28, 2015.

\bibitem[Hoyer et~al.(2023)Hoyer, Dai, Wang, and Van~Gool]{MIC}
Lukas Hoyer, Dengxin Dai, Haoran Wang, and Luc Van~Gool.
\newblock Mic: Masked image consistency for context-enhanced domain adaptation.
\newblock In \emph{CVPR}, pages 11721--11732, 2023.

\bibitem[Hu et~al.(2021)Hu, Wei, Hu, Ye, Cui, and Wang]{AEL}
Hanzhe Hu, Fangyun Wei, Han Hu, Qiwei Ye, Jinshi Cui, and Liwei Wang.
\newblock Semi-supervised semantic segmentation via adaptive equalization learning.
\newblock \emph{NeurIPS}, 34:\penalty0 22106--22118, 2021.

\bibitem[Huang et~al.(2023)Huang, Xie, Lin, Tong, Chen, Li, Wang, Huang, and Zheng]{Semicvt}
Huimin Huang, Shiao Xie, Lanfen Lin, Ruofeng Tong, Yen-Wei Chen, Yuexiang Li, Hong Wang, Yawen Huang, and Yefeng Zheng.
\newblock Semicvt: Semi-supervised convolutional vision transformer for semantic segmentation.
\newblock In \emph{CVPR}, pages 11340--11349, 2023.

\bibitem[Jin et~al.(2022)Jin, Wang, and Lin]{GTA}
Ying Jin, Jiaqi Wang, and Dahua Lin.
\newblock Semi-supervised semantic segmentation via gentle teaching assistant.
\newblock \emph{NeurIPS}, 35:\penalty0 2803--2816, 2022.

\bibitem[Ke et~al.(2022)Ke, Aviles-Rivero, Pandey, Reddy, and Sch{\"o}nlieb]{three_self_training}
Rihuan Ke, Angelica~I Aviles-Rivero, Saurabh Pandey, Saikumar Reddy, and Carola-Bibiane Sch{\"o}nlieb.
\newblock A three-stage self-training framework for semi-supervised semantic segmentation.
\newblock \emph{IEEE TIP}, 31:\penalty0 1805--1815, 2022.

\bibitem[Kim et~al.(2022)Kim, Min, Kim, Lee, Seo, Ryoo, and Kim]{kim2022conmatch}
Jiwon Kim, Youngjo Min, Daehwan Kim, Gyuseong Lee, Junyoung Seo, Kwangrok Ryoo, and Seungryong Kim.
\newblock Conmatch: Semi-supervised learning with confidence-guided consistency regularization.
\newblock In \emph{ECCV}, pages 674--690. Springer, 2022.

\bibitem[Kwon and Kwak(2022)]{ELN}
Donghyeon Kwon and Suha Kwak.
\newblock Semi-supervised semantic segmentation with error localization network.
\newblock In \emph{CVPR}, pages 9957--9967, 2022.

\bibitem[Lai et~al.(2021)Lai, Tian, Jiang, Liu, Zhao, Wang, and Jia]{CAC}
Xin Lai, Zhuotao Tian, Li Jiang, Shu Liu, Hengshuang Zhao, Liwei Wang, and Jiaya Jia.
\newblock Semi-supervised semantic segmentation with directional context-aware consistency.
\newblock In \emph{CVPR}, pages 1205--1214, 2021.

\bibitem[Laine and Aila(2016)]{Temporal}
Samuli Laine and Timo Aila.
\newblock Temporal ensembling for semi-supervised learning.
\newblock \emph{arXiv preprint arXiv:1610.02242}, 2016.

\bibitem[Li et~al.(2023{\natexlab{a}})Li, Purkait, Ajanthan, Abdolshah, Garg, Husain, Xu, Gould, Ouyang, and Van Den~Hengel]{DLG}
Peixia Li, Pulak Purkait, Thalaiyasingam Ajanthan, Majid Abdolshah, Ravi Garg, Hisham Husain, Chenchen Xu, Stephen Gould, Wanli Ouyang, and Anton Van Den~Hengel.
\newblock Semi-supervised semantic segmentation under label noise via diverse learning groups.
\newblock In \emph{CVPR}, pages 1229--1238, 2023{\natexlab{a}}.

\bibitem[Li et~al.(2023{\natexlab{b}})Li, He, Zhang, Zhang, Tan, Han, Ding, and Wang]{CFCG}
Shuo Li, Yue He, Weiming Zhang, Wei Zhang, Xiao Tan, Junyu Han, Errui Ding, and Jingdong Wang.
\newblock Cfcg: Semi-supervised semantic segmentation via cross-fusion and contour guidance supervision.
\newblock In \emph{CVPR}, pages 16348--16358, 2023{\natexlab{b}}.

\bibitem[Li et~al.(2024)Li, Jin, Wang, Wu, Liu, Tan, and Li]{SemiReward}
Siyuan Li, Weiyang Jin, Zedong Wang, Fang Wu, Zicheng Liu, Cheng Tan, and Stan~Z Li.
\newblock Semireward: A general reward model for semi-supervised learning.
\newblock In \emph{ICLR}, 2024.

\bibitem[Liang et~al.(2023)Liang, Wang, Miao, and Yang]{LOGICDIAG}
Chen Liang, Wenguan Wang, Jiaxu Miao, and Yi Yang.
\newblock Logic-induced diagnostic reasoning for semi-supervised semantic segmentation.
\newblock In \emph{CVPR}, pages 16197--16208, 2023.

\bibitem[Lin et~al.(2014)Lin, Maire, Belongie, Hays, Perona, Ramanan, Doll{\'a}r, and Zitnick]{coco}
Tsung-Yi Lin, Michael Maire, Serge Belongie, James Hays, Pietro Perona, Deva Ramanan, Piotr Doll{\'a}r, and C~Lawrence Zitnick.
\newblock Microsoft coco: Common objects in context.
\newblock In \emph{ECCV}, pages 740--755. Springer, 2014.

\bibitem[Liu et~al.(2022{\natexlab{a}})Liu, Jiang, Cao, Chen, Zhang, and Gui]{ljs1}
Jinshi Liu, Zhaohui Jiang, Ting Cao, Zhiwen Chen, Chaobo Zhang, and Weihua Gui.
\newblock Generated pseudo-labels guided by background skeletons for overcoming under-segmentation in overlapping particle objects.
\newblock \emph{TCSVT}, 33\penalty0 (6):\penalty0 2906--2919, 2022{\natexlab{a}}.

\bibitem[Liu et~al.(2024)Liu, Jiang, Gui, Chen, and Zhang]{ljs2}
Jinshi Liu, Zhaohui Jiang, Weihua Gui, Zhiwen Chen, and Chaobo Zhang.
\newblock Occlusion segmentation: Restore and segment invisible areas for particle objects.
\newblock \emph{IEEE Transactions on Automation Science and Engineering}, 2024.

\bibitem[Liu et~al.(2022{\natexlab{b}})Liu, Tian, Chen, Liu, Belagiannis, and Carneiro]{PS-MT}
Yuyuan Liu, Yu Tian, Yuanhong Chen, Fengbei Liu, Vasileios Belagiannis, and Gustavo Carneiro.
\newblock Perturbed and strict mean teachers for semi-supervised semantic segmentation.
\newblock In \emph{CVPR}, pages 4258--4267, 2022{\natexlab{b}}.

\bibitem[Ma et~al.(2023)Ma, Wang, Liu, Lin, and Li]{ESL}
Jie Ma, Chuan Wang, Yang Liu, Liang Lin, and Guanbin Li.
\newblock Enhanced soft label for semi-supervised semantic segmentation.
\newblock In \emph{CVPR}, pages 1185--1195, 2023.

\bibitem[Mai et~al.(2024)Mai, Sun, Zhang, and Wu]{rankmatch}
Huayu Mai, Rui Sun, Tianzhu Zhang, and Feng Wu.
\newblock Rankmatch: Exploring the better consistency regularization for semi-supervised semantic segmentation.
\newblock In \emph{CVPR}, pages 3391--3401, 2024.

\bibitem[Mittal et~al.(2019)Mittal, Tatarchenko, and Brox]{mittal2019semi}
Sudhanshu Mittal, Maxim Tatarchenko, and Thomas Brox.
\newblock Semi-supervised semantic segmentation with high-and low-level consistency.
\newblock \emph{TEEE TPAMI}, 43\penalty0 (4):\penalty0 1369--1379, 2019.

\bibitem[Miyato et~al.(2018)Miyato, Maeda, Koyama, and Ishii]{vat}
Takeru Miyato, Shin-ichi Maeda, Masanori Koyama, and Shin Ishii.
\newblock Virtual adversarial training: a regularization method for supervised and semi-supervised learning.
\newblock \emph{IEEE TPAMI}, 41\penalty0 (8):\penalty0 1979--1993, 2018.

\bibitem[Olsson et~al.(2008)Olsson, Eriksson, and Kahl]{relaxation}
Carl Olsson, Anders~P Eriksson, and Fredrik Kahl.
\newblock Improved spectral relaxation methods for binary quadratic optimization problems.
\newblock \emph{Computer Vision and Image Understanding}, 112\penalty0 (1):\penalty0 3--13, 2008.

\bibitem[Qiao et~al.(2023)Qiao, Wei, Wang, Wang, Song, Xu, Ji, Liu, and Chen]{FPL}
Pengchong Qiao, Zhidan Wei, Yu Wang, Zhennan Wang, Guoli Song, Fan Xu, Xiangyang Ji, Chang Liu, and Jie Chen.
\newblock Fuzzy positive learning for semi-supervised semantic segmentation.
\newblock In \emph{CVPR}, pages 15465--15474, 2023.

\bibitem[Rizve et~al.(2021)Rizve, Duarte, Rawat, and Shah]{defense}
Mamshad~Nayeem Rizve, Kevin Duarte, Yogesh~S Rawat, and Mubarak Shah.
\newblock In defense of pseudo-labeling: An uncertainty-aware pseudo-label selection framework for semi-supervised learning.
\newblock \emph{arXiv preprint arXiv:2101.06329}, 2021.

\bibitem[Sajjadi et~al.(2016)Sajjadi, Javanmardi, and Tasdizen]{sregularization}
Mehdi Sajjadi, Mehran Javanmardi, and Tolga Tasdizen.
\newblock Regularization with stochastic transformations and perturbations for deep semi-supervised learning.
\newblock \emph{NeurIPS}, 29, 2016.

\bibitem[Shin et~al.(2021)Shin, Xie, and Albanie]{shin2021all}
Gyungin Shin, Weidi Xie, and Samuel Albanie.
\newblock All you need are a few pixels: semantic segmentation with pixelpick.
\newblock In \emph{ICCV}, pages 1687--1697, 2021.

\bibitem[Shrivastava et~al.(2016)Shrivastava, Gupta, and Girshick]{OHEM}
Abhinav Shrivastava, Abhinav Gupta, and Ross Girshick.
\newblock Training region-based object detectors with online hard example mining.
\newblock In \emph{CVPR}, pages 761--769, 2016.

\bibitem[Sohn et~al.(2020)Sohn, Berthelot, Carlini, Zhang, Zhang, Raffel, Cubuk, Kurakin, and Li]{fixmatch}
Kihyuk Sohn, David Berthelot, Nicholas Carlini, Zizhao Zhang, Han Zhang, Colin~A Raffel, Ekin~Dogus Cubuk, Alexey Kurakin, and Chun-Liang Li.
\newblock Fixmatch: Simplifying semi-supervised learning with consistency and confidence.
\newblock \emph{NeurIPS}, 33:\penalty0 596--608, 2020.

\bibitem[Sun et~al.(2024{\natexlab{a}})Sun, Yang, Zhang, Cheng, and Hou]{CorrMatch}
Boyuan Sun, Yuqi Yang, Le Zhang, Ming-Ming Cheng, and Qibin Hou.
\newblock Corrmatch: Label propagation via correlation matching for semi-supervised semantic segmentation.
\newblock In \emph{CVPR}, pages 3097--3107, 2024{\natexlab{a}}.

\bibitem[Sun et~al.(2024{\natexlab{b}})Sun, Mai, Zhang, and Wu]{DAW}
Rui Sun, Huayu Mai, Tianzhu Zhang, and Feng Wu.
\newblock Daw: exploring the better weighting function for semi-supervised semantic segmentation.
\newblock \emph{NeurIPS}, 36, 2024{\natexlab{b}}.

\bibitem[Tarvainen and Valpola(2017)]{MT}
Antti Tarvainen and Harri Valpola.
\newblock Mean teachers are better role models: Weight-averaged consistency targets improve semi-supervised deep learning results.
\newblock \emph{NeurIPS}, 30, 2017.

\bibitem[Tian(1994)]{kf}
GQ Tian.
\newblock Generalized kkm theorems, minimax inequalities, and their applications.
\newblock \emph{Journal of Optimization Theory and Applications}, 83:\penalty0 375--389, 1994.

\bibitem[Wang et~al.(2024{\natexlab{a}})Wang, Zhang, Li, and Li]{allspark}
Haonan Wang, Qixiang Zhang, Yi Li, and Xiaomeng Li.
\newblock Allspark: Reborn labeled features from unlabeled in transformer for semi-supervised semantic segmentation.
\newblock In \emph{CVPR}, pages 3627--3636, 2024{\natexlab{a}}.

\bibitem[Wang et~al.(2022{\natexlab{a}})Wang, Nie, Fang, Han, Wu, Wang, Lin, Zhou, and Li]{Double}
Kuo Wang, Yuxiang Nie, Chaowei Fang, Chengzhi Han, Xuewen Wu, Xiaohui Wang, Liang Lin, Fan Zhou, and Guanbin Li.
\newblock Double-check soft teacher for semi-supervised object detection.
\newblock In \emph{IJCAI}, pages 1430--1436, 2022{\natexlab{a}}.

\bibitem[Wang et~al.(2020)Wang, Long, Wang, and Jordan]{PC}
Ximei Wang, Mingsheng Long, Jianmin Wang, and Michael Jordan.
\newblock Transferable calibration with lower bias and variance in domain adaptation.
\newblock \emph{Advances in Neural Information Processing Systems}, 33:\penalty0 19212--19223, 2020.

\bibitem[Wang et~al.(2023{\natexlab{a}})Wang, Zhang, Yu, and Xiao]{DGCL}
Xiaoyang Wang, Bingfeng Zhang, Limin Yu, and Jimin Xiao.
\newblock Hunting sparsity: Density-guided contrastive learning for semi-supervised semantic segmentation.
\newblock In \emph{CVPR}, pages 3114--3123, 2023{\natexlab{a}}.

\bibitem[Wang et~al.(2024{\natexlab{b}})Wang, Bai, Yu, Zhao, and Xiao]{DDFP}
Xiaoyang Wang, Huihui Bai, Limin Yu, Yao Zhao, and Jimin Xiao.
\newblock Towards the uncharted: Density-descending feature perturbation for semi-supervised semantic segmentation.
\newblock In \emph{CVPR}, pages 3303--3312, 2024{\natexlab{b}}.

\bibitem[Wang et~al.(2022{\natexlab{b}})Wang, Chen, Fan, Sun, Tao, Hou, Wang, Yang, Zhou, Guo, et~al.]{Usb}
Yidong Wang, Hao Chen, Yue Fan, Wang Sun, Ran Tao, Wenxin Hou, Renjie Wang, Linyi Yang, Zhi Zhou, Lan-Zhe Guo, et~al.
\newblock Usb: A unified semi-supervised learning benchmark for classification.
\newblock \emph{NeurIPS}, 35:\penalty0 3938--3961, 2022{\natexlab{b}}.

\bibitem[Wang et~al.(2022{\natexlab{c}})Wang, Wang, Shen, Fei, Li, Jin, Wu, Zhao, and Le]{U2PL}
Yuchao Wang, Haochen Wang, Yujun Shen, Jingjing Fei, Wei Li, Guoqiang Jin, Liwei Wu, Rui Zhao, and Xinyi Le.
\newblock Semi-supervised semantic segmentation using unreliable pseudo-labels.
\newblock In \emph{CVPR}, pages 4248--4257, 2022{\natexlab{c}}.

\bibitem[Wang et~al.(2023{\natexlab{b}})Wang, Chen, Heng, Hou, Fan, Wu, Wang, Savvides, Shinozaki, Raj, et~al.]{freematch}
Yidong Wang, Hao Chen, Qiang Heng, Wenxin Hou, Yue Fan, Zhen Wu, Jindong Wang, Marios Savvides, Takahiro Shinozaki, Bhiksha Raj, et~al.
\newblock Freematch: Self-adaptive thresholding for semi-supervised learning.
\newblock In \emph{ICLR}, 2023{\natexlab{b}}.

\bibitem[Wang et~al.(2023{\natexlab{c}})Wang, Zhao, Xing, Xu, Kong, and Zhou]{CCVC}
Zicheng Wang, Zhen Zhao, Xiaoxia Xing, Dong Xu, Xiangyu Kong, and Luping Zhou.
\newblock Conflict-based cross-view consistency for semi-supervised semantic segmentation.
\newblock In \emph{CVPR}, pages 19585--19595, 2023{\natexlab{c}}.

\bibitem[Wang et~al.(2025)Wang, Chen, Liu, Zhao, Zhu, and Wu]{Seg3}
Zhenyan Wang, Zhenxue Chen, Chengyun Liu, Yaping Zhao, Xinming Zhu, and QM~Jonathan Wu.
\newblock Diversity augmentation and multi-fuzzy label for semi-supervised semantic segmentation.
\newblock \emph{Neurocomputing}, page 129681, 2025.

\bibitem[Xie et~al.(2020{\natexlab{a}})Xie, Dai, Hovy, Luong, and Le]{Unsupervised}
Qizhe Xie, Zihang Dai, Eduard Hovy, Thang Luong, and Quoc Le.
\newblock Unsupervised data augmentation for consistency training.
\newblock \emph{NeurIPS}, 33:\penalty0 6256--6268, 2020{\natexlab{a}}.

\bibitem[Xie et~al.(2020{\natexlab{b}})Xie, Luong, Hovy, and Le]{xie2020self}
Qizhe Xie, Minh-Thang Luong, Eduard Hovy, and Quoc~V Le.
\newblock Self-training with noisy student improves imagenet classification.
\newblock In \emph{CVPR}, pages 10687--10698, 2020{\natexlab{b}}.

\bibitem[Xu et~al.(2022)Xu, Liu, Bian, and Yang]{xu2022semi}
Haiming Xu, Lingqiao Liu, Qiuchen Bian, and Zhen Yang.
\newblock Semi-supervised semantic segmentation with prototype-based consistency regularization.
\newblock \emph{NeurIPS}, 35:\penalty0 26007--26020, 2022.

\bibitem[Xu et~al.(2021)Xu, Shang, Ye, Qian, Li, Sun, Li, and Jin]{Dash}
Yi Xu, Lei Shang, Jinxing Ye, Qi Qian, Yu-Feng Li, Baigui Sun, Hao Li, and Rong Jin.
\newblock Dash: Semi-supervised learning with dynamic thresholding.
\newblock In \emph{ICML}, pages 11525--11536. PMLR, 2021.

\bibitem[Yang et~al.(2022)Yang, Zhuo, Qi, Shi, and Gao]{ST++}
Lihe Yang, Wei Zhuo, Lei Qi, Yinghuan Shi, and Yang Gao.
\newblock St++: Make self-trainingwork better for semi-supervised semantic segmentation.
\newblock In \emph{CVPR}, pages 4258--4267, 2022.

\bibitem[Yang et~al.(2023)Yang, Qi, Feng, Zhang, and Shi]{UniMatch}
Lihe Yang, Lei Qi, Litong Feng, Wayne Zhang, and Yinghuan Shi.
\newblock Revisiting weak-to-strong consistency in semi-supervised semantic segmentation.
\newblock In \emph{CVPR}, pages 7236--7246, 2023.

\bibitem[Yang et~al.(2025)Yang, Zhao, and Zhao]{Unimatchv2}
Lihe Yang, Zhen Zhao, and Hengshuang Zhao.
\newblock Unimatch v2: Pushing the limit of semi-supervised semantic segmentation.
\newblock \emph{IEEE TPAMI}, 2025.

\bibitem[Yun et~al.(2019)Yun, Han, Chun, Oh, Yoo, and Choe]{CutMix}
Sangdoo Yun, Dongyoon Han, Sanghyuk Chun, Seong~Joon Oh, Youngjoon Yoo, and Junsuk Choe.
\newblock Cutmix: Regularization strategy to train strong classifiers with localizable features.
\newblock In \emph{ICCV}, pages 6022--6031, 2019.

\bibitem[Zhang et~al.(2021)Zhang, Wang, Hou, Wu, Wang, Okumura, and Shinozaki]{flexmatch}
Bowen Zhang, Yidong Wang, Wenxin Hou, Hao Wu, Jindong Wang, Manabu Okumura, and Takahiro Shinozaki.
\newblock Flexmatch: Boosting semi-supervised learning with curriculum pseudo labeling.
\newblock \emph{NeurIPS}, 34:\penalty0 18408--18419, 2021.

\bibitem[Zhao et~al.(2023{\natexlab{a}})Zhao, Qi, Wang, Wang, Wu, Mao, and Zhang]{RCPS}
Xiangyu Zhao, Zengxin Qi, Sheng Wang, Qian Wang, Xuehai Wu, Ying Mao, and Lichi Zhang.
\newblock Rcps: Rectified contrastive pseudo supervision for semi-supervised medical image segmentation.
\newblock \emph{IEEE Journal of Biomedical and Health Informatics}, 2023{\natexlab{a}}.

\bibitem[Zhao et~al.(2023{\natexlab{b}})Zhao, Yang, Long, Pi, Zhou, and Wang]{augseg}
Zhen Zhao, Lihe Yang, Sifan Long, Jimin Pi, Luping Zhou, and Jingdong Wang.
\newblock Augmentation matters: A simple-yet-effective approach to semi-supervised semantic segmentation.
\newblock In \emph{CVPR}, pages 11350--11359, 2023{\natexlab{b}}.

\bibitem[Zoph et~al.(2020)Zoph, Ghiasi, Lin, Cui, Liu, Cubuk, and Le]{zoph2020rethinking}
Barret Zoph, Golnaz Ghiasi, Tsung-Yi Lin, Yin Cui, Hanxiao Liu, Ekin~Dogus Cubuk, and Quoc Le.
\newblock Rethinking pre-training and self-training.
\newblock \emph{NeurIPS}, 33:\penalty0 3833--3845, 2020.

\bibitem[Zou et~al.(2020)Zou, Zhang, Zhang, Li, Bian, Huang, and Pfister]{pseudoseg}
Yuliang Zou, Zizhao Zhang, Han Zhang, Chun-Liang Li, Xiao Bian, Jia-Bin Huang, and Tomas Pfister.
\newblock Pseudoseg: Designing pseudo labels for semantic segmentation.
\newblock In \emph{ICLR}, 2020.

\end{thebibliography}
}

\maketitlesupplementary
\appendix
\section{Overview}

In the supplementary material for CSL, we provide a proof of \cref{eq:5} (Sec.\ref{sec:proof}),  give a low-complexity implementation of CSL and pseudo-code (Sec.\ref{sec:algorithm}), extend the implementation details and experimental comparison (Sec.\ref{sec:add_experiment}), supplements the analysis of residual dispersion(Sec.\ref{sec:residual}), discuss the potential limitations (Sec.\ref{sec:limit}), and gives more visual results (Sec.\ref{sec:visual}).

\section{Proof of Equation \ref{eq:5}} 
\label{sec:proof}
For the pseudo-label selection problem in semi-supervised semantic segmentation, it can be formulated as:
\[
    \mathop{\max}\limits_{S}Tr\left(S^T \Phi^T \Phi S\right), s.t.S\in\{0,1\}^{HW\times2},
\]
where $S$ is the binary selection matrix subject to  $\sum_{c} S_{n,c} = 1$ with $c \in \{1,2\}$ indicating class indices, and $Tr(\cdot)$ denotes the matrix trace. Each column of $\Phi=[\mathbf{h}_1,\mathbf{h}_2,...,\mathbf{h}_{HW}]$ represents the feature $\mathbf{h}_n \in \mathbb{R}^{2}$ of pixel $n$.

\noindent\textit{Proof.} Without loss of generality, the essential goal of pseudo-label selection is to assign pixel classes via $S$ such that the selected pseudo-labels are optimal under a certain metric:
\[
\mathcal{L}(S)=\sum_{n=1}^{HW}\varphi\left(\mathbf{z}_n,S_{n,:}\right),
\]
where $\varphi(\cdot,\cdot)$ is a function that measures the potential risk of assigning pixel $n$ to a specific class , with $\mathbf{z}_n$ representing the pixel's feature. Considering the potential natural separation exhibited in \cref{fig:1}, we employ intra-class consistency as the metric, which measures the distance between the pixel feature vector $\mathbf{z}_n=\mathbf{h}_n$ and the mean feature vector $\mathbf{\mu}_c$ of its assigned class $c$, which can be expressed as
\[
\varphi\left(\mathbf{z}_n,S_{n,:}\right)=
\sum_{c=1}^2 S_{n,c} \|\mathbf{h}_n(c) - \mathbf{\mu}_c\|^2,
\]
thus $\mathcal{L}(S)$ can be reformulated as
\[
\mathcal{L}(S) = \|\Phi - S (S^T S)^{-1} S^T \Phi\|_F^2,
\]
let \(\mathbf{P} = S (S^T S)^{-1} S^T\), the $\mathcal{L}(S)$ simplifies to
\[
\mathcal{L}(S) = \|\Phi - \mathbf{P} \Phi\|_F^2.
\]
Using the Frobenius norm identity
\[
\mathcal{L}(S) = \|\Phi\|_F^2 - 2 \operatorname{Tr}(\Phi^T \mathbf{P} \Phi) + \|\mathbf{P} \Phi\|_F^2,
\]
since \(\mathbf{P}^2 = \mathbf{P}\) (idempotence of projection matrices), thus
\[
\mathcal{L}(S) = \|\Phi\|_F^2 - \operatorname{Tr}(\Phi^T \mathbf{P} \Phi).
\]
Minimizing \(\mathcal{L}(S)\) is equivalent to maximizing \(\operatorname{Tr}(\Phi^T \mathbf{P} \Phi)\). Substituting \(\mathbf{P}\):
\[
\operatorname{Tr}(\Phi^T \mathbf{P} \Phi) = \operatorname{Tr}(\Phi^T S (S^T S)^{-1} S^T \Phi),
\]
for binary classification, \(S^T S = \text{diag}(n_1, n_2)\), where \(n_1\) and \(n_2\) are the number of pixels in each class. Thus:
\[
\operatorname{Tr}(\Phi^T \mathbf{P} \Phi) = \operatorname{Tr}(S^T \Phi \Phi^T S) \cdot (n_1^{-1} + n_2^{-1}),
\]
ignoring the normalization constant, this reduces to:
\[
\mathop{\max}\limits_{S} \operatorname{Tr}(S^T \Phi^T \Phi S).
\]

\section{Algorithm}
\label{sec:algorithm}

\subsection{Low-complexity Implementation of Prediction Convex Optimization Separation}
In \cref{eq:6}, we used the eigenvectors $u_i$ of $\Phi^T \Phi$. however, obtaining $u_i$ through $\Phi^T \Phi$ is computationally expensive, especially when $\Phi \in \mathbb{R}^{2 \times HW}$ has a large number of columns, where $HW \gg 2$. so we provide an efficient alternative with Singular Value Decomposition (SVD).

The SVD of \(\Phi\) is given by:
\[
\Phi = \mathbf{U} \mathbf{\Sigma} \mathbf{V}^T,
\]
where \(\mathbf{U} \in \mathbb{R}^{2 \times 2}\) and \(\mathbf{V} \in \mathbb{R}^{HW \times HW}\)  are orthogonal matrix, \(\mathbf{\Sigma} \in \mathbb{R}^{2 \times HW}\) is a diagonal matrix containing the singular values \(\sigma_i\).

Substituting the SVD of \(\Phi\),
\[
\Phi^T \Phi = (\mathbf{U} \mathbf{\Sigma} \mathbf{V}^T)^T (\mathbf{U} \mathbf{\Sigma} \mathbf{V}^T),
\]
which simplifies to
\[
\Phi^T \Phi = \mathbf{V} \mathbf{\Sigma}^T \mathbf{U}^T \mathbf{U} \mathbf{\Sigma} \mathbf{V}^T.
\]
Using the orthogonality of \(\mathbf{U}\) (\(\mathbf{U}^T \mathbf{U} = \mathbf{I}\)), 
\[
\Phi^T \Phi = \mathbf{V} \mathbf{\Sigma}^T \mathbf{\Sigma} \mathbf{V}^T.
\]
Let \(\mathbf{\Lambda} = \mathbf{\Sigma}^T \mathbf{\Sigma} = \text{diag}(\sigma_1^2, \sigma_2^2)\), where \(\sigma_i^2\) are the squared singular values. Then
\[
\Phi^T \Phi = \mathbf{V} \mathbf{\Lambda} \mathbf{V}^T.
\]
This shows that \(\mathbf{V}\) contains the eigenvectors of \(\Phi^T \Phi\), \(\mathbf{\Lambda}\) contains the eigenvalues of \(\Phi^T \Phi\).

Computing the eigenvectors of \(\Phi^T \Phi\) via direct eigen-decomposition requires explicitly forming \(\Phi^T \Phi\), which has a cost of \(\mathcal{O}(HW^2)\) for matrix multiplication and \(\mathcal{O}(HW^3)\) for eigen-decomposition. In contrast, computing the SVD of \(\Phi\) has a cost of \(\mathcal{O}(HW^2)\).

\subsection{Pseudocode for Optimization Separation} 

Algorithm \ref{alg:1} provides the pseudocode for Prediction Convex Optimization Separation. Using a convex optimization strategy based on confidence distribution, CSL effectively excludes a large number of high-confidence false predictions in pseudo-label selection to improve the performance in semi-supervised semantic segmentation tasks.

\definecolor{commentgreen}{RGB}{0, 128, 0} 
\newcommand{\greencomment}[1]{\hspace{1em}\textcolor{commentgreen}{\footnotesize \texttt{\# #1}}}
\newcommand{\comment}[1]{\textcolor{commentgreen}{\footnotesize \texttt{\# #1}}}
\newcommand{\DEF}[1]{\STATE\footnotesize \texttt{#1}}
\newcommand{\INDSTATE}[2][1]{\STATE\hspace{#1\algorithmicindent}\footnotesize \texttt{#2}}

\begin{algorithm}[H]
\caption{Pseudocode of Prediction Convex Optimization Separation in a PyTorch-like style.}
\label{alg:1}
\greencomment{pmax:\hspace{0.5em}Pixel-level maximum confidence}\vspace{-0.3em}

\greencomment{vn:\hspace{0.5em}Pixel-level residual dispersion}\vspace{-0.3em}
\\\vspace{-0.7em}

\begin{algorithmic} 

\DEF {def PCOS(pmax, vn): }

\greencomment{combine pmax and vn into a feature matrix $\Phi$}

\INDSTATE {$\Phi$ = stack([pmax, vn], axis=1).T}

\greencomment{extract the top two eigenvectors from $\Phi$}

\INDSTATE {U, Sigma, VT = svd($\Phi$)}

\INDSTATE {eig\_vectors = VT[:, :2] }

\greencomment{constructing the optimal selection matrix}

\INDSTATE {S = argmax(abs(eig\_vectors), axis=1)}

\greencomment{calculate stats for each class}

\INDSTATE {stats = [($\Phi$[S == c].mean(dim=1),\\ 
\quad \quad \quad $\Phi$[S == c].std(dim=1)) for c in range(2)]}

\greencomment{select the reliable class}

\INDSTATE {mu, sigma = max(stats, key=lambda x:x[0])}

\greencomment{smooth loss weight}

\INDSTATE {weight = exp(-(($\Phi$-mu)/(8*sigma))**2)}

\INDSTATE{weight = weight.prod(dim=0)}

\greencomment{preserving reliable prediction weights}

\INDSTATE {weight[($\Phi$[0, :]\text{\textgreater}mu[0])$\|$($\Phi$[1, :]\text{\textgreater}mu[1])] = 1 }

\INDSTATE {return weight}

    \end{algorithmic}
\end{algorithm}

\subsection{Pseudocode for Trusted Mask Perturbation}

In \cref{subsec:3.4} of our paper, we propose the Trusted Mask Perturbation Strategy. The core idea of this method is to enhance the mutual information between low-confidence regions and pseudo-labels to strengthen the fitting of these regions. Specifically, we use high-confidence predictions from weakly augmented outputs as pseudo-labels but randomly discard their image content, while the image content of low-confidence predictions is entirely preserved. This forces the network to infer the classes of high-confidence regions based on the image content of low-confidence regions, thereby compensating for the underfitting in low-confidence areas. To clarify things, we present the pseudocode of the threshold updating strategy in a PyTorch-like style.

\begin{algorithm}[H]
\caption{Pseudocode of Trusted Mask Perturbation in a PyTorch-like style.}
\label{alg:2}
\greencomment{x\_w:\hspace{0.2em}Image with weak augmentation perturbation}\vspace{-0.3em}

\greencomment{image\_size:\hspace{0.5em}The length or width of the image}\vspace{-0.3em}

\greencomment{block\_size:\hspace{0.5em}The masking patch size}\vspace{-0.3em}

\greencomment{masking\_rate:\hspace{0.5em}The masking pixel ratio}\vspace{-0.3em}

\greencomment{f:\hspace{0.5em}segmentation network}\vspace{-0.3em}
\\\vspace{-0.7em}

\begin{algorithmic} 

\DEF {pred\_w = f(x\_w)}

\DEF {mask\_w = pred\_w.argmax(dim=1).detach()}

\comment{compute weights using PCOS on the projection}

\DEF {weight = PCOS(Projection(pred\_w))}

\comment{create a reliability mask}

\DEF {reli\_mask = (weight == 1)}

\comment{gain patch-based perturbation mask (\cref{eq:9})}

\DEF {mask\_size = img\_size // block\_size}

\DEF {cover\_mask = (rand(mask\_size, mask\_size) \text{\textless}\\
\quad \quad \quad \quad \quad \quad \quad masking\_rate).float()}

\DEF {cover\_mask = interpolate(cover\_mask, \\
\quad \quad \quad \quad \quad \quad \quad size=img\_size, mode='nearest')}

\comment{perturbation only for reliable predictions}

\DEF {cover\_mask =  cover\_mask $\&$ reli\_mask}

\comment {constructing perturbed images}
\DEF {x\_m = x\_s.clone()}

\DEF {x\_m[cover\_mask == 1] = 0}

\DEF {pred\_m = f(x\_m)}

\comment {calculated loss}
\DEF {criterion = CrossEntropyLoss()}

\DEF {loss\_m = criterion(pred\_m, mask\_w)}

    \end{algorithmic}
\end{algorithm}

\section{Extensive Experiment Details and Results}
\label{sec:add_experiment}  

\subsection{More Implementation Details}

Following prior works~\cite{UniMatch,LOGICDIAG} , we employ random scaling between [0.5, 2.0], cropping, and flipping as weak augmentations. We combine ColorJitter, random grayscale, Gaussian blur, and CutMix~\cite{CutMix} for strong augmentations. Weak augmentations and a modified strong augmentation (without random grayscale and Gaussian blur) are applied before Trusted Mask Perturbation.  Additionally, we incorporate 50\% random channel dropout as feature perturbations to encourage robust feature representations as in previous works~\cite{UniMatch,CorrMatch,DAW} . For computational efficiency, mixed-precision training based on BrainFloat16 is utilized.

\subsection{More Experimental Results}

\noindent\textbf{Different loss weights.} In \cref{tab:7}, we investigate the impact of different loss weightings on segmentation accuracy. The imbalance between consistency loss and masking loss significantly affects model performance. Moreover, assigning excessively high or low loss weights to the unlabeled data also leads to a degradation in performance. The results indicate that the optimal performance is achieved when $\left[\lambda_1, \lambda_2\right]$ are set to [0.5, 0.5].

\begin{table}[t]
  \centering
  \begin{tabularx}{\columnwidth}{*{2}{c}|*{3}{>{\centering\arraybackslash}X}}
    \toprule[1.1pt]
    $\lambda_1$ & $\lambda_2$ & 1/16(92) & 1/8(183) & 1/2(732) \\
        \midrule
    0.50 & 0.50 & \textbf{76.8} & \textbf{79.6} & \textbf{80.9} \\
    0.75 & 0.25 & 75.6          & 78.1          & 79.7          \\
    0.25 & 0.75 & 75.2          & 77.6          & 79.2          \\
    0.30 & 0.30 & 76.3          & 79.1          & 80.2          \\
    0.70 & 0,70 & 75.7          & 78.4          & 79.8          \\
    1.00 & 1.00 & 72.5          & 76.6          & 77.6          \\
    \bottomrule[1.1pt]
  \end{tabularx}
  \caption{Impact of different loss weights, evaluated on the original PASCAL VOC 2012 with a crop size of 321.}
  \label{tab:7}
\end{table}

\begin{table}[t]
  \centering
  \begin{tabularx}{\columnwidth}{*{1}{c}|*{5}{>{\centering\arraybackslash}X}}
    \toprule[1.1pt]
    Method & 1/16 & 1/8 & 1/4 & 1/2 & Full \\
    \midrule
    CSL  & \textbf{76.8} & \textbf{79.6} & \textbf{80.3} & \textbf{80.9} & \textbf{82.3}\\
    Class-specific CSL  & 74.6 & 78.3 & 79.1 & 79.7 & 80.6  \\
    \bottomrule[1.1pt]
  \end{tabularx}
  \caption{Comparison of CSL and class-specific strategy, evaluated on the original PASCAL VOC 2012 with a crop size of 321.}
  \label{tab:8}
\end{table}

\begin{table}[t]
\centering
\setlength{\tabcolsep}{2.4pt}
\small
\begin{tabular}{l|c|ccccccc|c}
\hline
Metrics & & plane & bicyc & bus & car & chair & perso & sofa& mean\\
\hline
Sampling & \parbox[t]{2mm}{\multirow{3}{*}{\rotatebox[origin=c]{90}{Base}}}
& \textbf{96.2} & \textbf{93.1} & \textbf{98.0} & \textbf{93.2} & \textbf{82.0} & \textbf{91.0} & \textbf{78.6} & \textbf{91.0} \\
Accuracy &  
& 90.8 & 75.4 & 90.1 & 85.4 & 33.4 & 83.7 & 50.2 & 82.4 \\
Recall &  
& 96.6 & 97.6 & 93.2 & 89.5 & \textbf{81.8} & 87.5 & 86.7 & 93.6 \\
\hline
Sampling & \parbox[t]{2mm}{\multirow{3}{*}{\rotatebox[origin=c]{90}{CSL}}} 
& 92.3 & 85.7 & 94.1 & 89.5 & 63.1 & 85.6 & 71.2& 87.6 \\
Accuracy &  
& \textbf{95.1} & \textbf{82.4} & \textbf{96.4} & \textbf{91.7} & \textbf{42.5} & \textbf{89.1} & \textbf{60.4} & \textbf{86.5} \\
Recall &  
& \textbf{97.1} & \textbf{98.2} & \textbf{95.8} & \textbf{92.3} & 80.1 & \textbf{87.7} & \textbf{94.5} & \textbf{94.6} \\
\toprule
\end{tabular}
\caption{Pseudo-Label sampling rate, accuracy and recall comparison under PASCAL original 1/4 Splits with class-wise metrics.}
\label{tab:12}
\end{table}

\noindent\textbf{A:}We conducted experiments comparing threshold-based methods and CSL on identical model predictions. As shown in \cref{tab:12}, CSL achieves consistent accuracy improvements (+4.1\%), while maintaining competitive recall (+1.0\%).

\noindent\textbf{Class-specific prediction selection.} In semi-supervised semantic segmentation, due to the long-tailed distribution of datasets, the confidence distributions of predictions vary significantly across classes. This suggests that employing a class-specific convex optimization strategy could potentially yield performance gains. To analyze this, we reconstruct the feature matrix $\Phi$ into class-specific feature spaces for Prediction Convex Optimization Separation:
\begin{equation}
    \Phi_k = \{h_n \mid h_n \in \Phi, k_n^\ast = k\}
    \label{eq:13}
\end{equation}
where $\Phi_c$ is Class-specific feature matrix for class $k$ and $k_n^\ast$ is the predicted class label for pixel $n$.

We conducted ablation experiments presented in \cref{tab:8}, where it can be observed that using class-specific schemes results in significant performance degradation. This may be attributed to the fact that most classes have too few pixel samples within instances to maintain an effective convex optimization strategy.

\noindent\textbf{Augmentations of Trusted Mask Perturbations.} In \cref{tab:9}, we evaluate the impact of various augmentations on Trusted Mask Perturbations. Adding CutMix leads to significant performance degradation, as it introduces misleading contextual information by directly stitching image patches while the trusted masking mechanism forces the model to learn contextual relationships, resulting in negative effects. Similarly, random grayscale enhances the sample by reducing color diversity, which conflicts with the masking mechanism and results in substantial information loss. Therefore, we adopt enhancements that exclude CutMix and random grayscale as additional Augmentations to the masking strategy.

\begin{table}[t]
  \centering
  \begin{tabularx}{\columnwidth}{c|*{5}{>{\centering\arraybackslash}X}}
    \toprule[1.1pt]
    dataset & 92 & 183 & 366 & 732 & 1464 \\
    \midrule
    $D_l$          & 73.2  & 77.1  & 78.8  & 79.4  & {80.3} \\
    $D_u$          & \textbf{76.8}  & \textbf{79.6}  & \textbf{80.3}  & \textbf{80.9}  & \textbf{82.3} \\
    $D_l \cup D_u$ & 75.6  & 78.4  & 79.6  & 80.1  & {81.7} \\
    \bottomrule[1.1pt]
  \end{tabularx}
  \caption{Ablation study of masking datasets under 1/2 splits. For the labeled dataset, predictions are treated as reliable predictions.}
  \label{tab:5}
\end{table}

\noindent\textbf{Selection of Masking Datasets.} To evaluate the impact of applying the mask perturbation to different subsets: labeled images, unlabelled images, and the combination of both, experiments were performed under different splits, as detailed in \cref{tab:5}. Results show applying masking to labeled data or the combination leads to progressively severe performance degradation as the number of labeled data samples decreases. This phenomenon can be attributed to the network learning contextual relationships present only in the labeled data instances and applying them to unlabeled data.

\begin{table}[t]
  \centering
  \begin{tabularx}{\columnwidth}{*{1}{l}|*{2}{>{\centering\arraybackslash}X}}
    \toprule[1.1pt]
    Method  & 1/8  & 1/2 \\
    \midrule
    $A^\omega$ only & 78.3 & 79.1 \\
    $A^\omega \& A^s$ & 78.7 & 79.5 \\
    $A^\omega \& A^s$ W/O Cutmix & 79.1 & 79.8 \\
    $A^\omega \& A^s$ W/O grayscale & 79.0 & 80.4 \\
    $A^\omega \& A^s$ W/O (Cutmix \& grayscale) & \textbf{79.6} & \textbf{80.9} \\
    \bottomrule[1.1pt]
  \end{tabularx}
  \caption{Comparison of augmentations strategies for TMP, evaluated on the original PASCAL VOC 2012 with a crop size of 321.}
  \label{tab:9}
\end{table}

\begin{table}[t]
  \centering
  \small
  \setlength{\tabcolsep}{3pt}
  \begin{tabularx}{\columnwidth}{@{}>{\hsize=1.55\hsize\raggedright\arraybackslash}X 
                                  >{\hsize=1.3\hsize\raggedright\arraybackslash}X 
                                  *{5}{>{\hsize=0.8\hsize\centering\arraybackslash}X}@{}}
    \specialrule{1.1pt}{0pt}{0pt}
    Method & Encoder & 1/16 & 1/8 & 1/4 & 1/2 & full\\
    \specialrule{0.1pt}{0pt}{0.1pt}
    UniMatchV2 & DINOv2 & 79.0 & 85.5 & 85.9 & 86.7 & 87.8 \\
    Ours & DINOv2 & \textbf{80.2} & \textbf{85.8} & \textbf{86.3} & \textbf{87.4} & \textbf{88.1} \\
    \specialrule{0.1pt}{0pt}{0.1pt}
    AllSpark & SegFormer & 76.1 & 78.4 & 79.8 & 80.8 & 82.1 \\
    Ours & SegFormer & \textbf{77.4} & \textbf{80.2} & \textbf{81.5} & \textbf{83.5} & \textbf{85.3} \\
    \bottomrule[1.1pt]
  \end{tabularx}
  \caption{Influence of different network architectures, evaluated on the original PASCAL VOC 2012 with a crop size of 513.}
  \label{tab:10}
\end{table}

\noindent\textbf{Different network Architectures.} Considering that different encoders may exhibit varying degrees of overconfidence in their representations and utilize contextual relationships through distinct mechanisms, we supplement additional experiments with diverse network architectures in \cref{tab:10}. Specifically, DINOv2-S adopts the hyperparameters from Unimatchv2 \cite{Unimatchv2}, while SegFormer-B5\cite{allspark} shares the training configuration described in \cref{subsec:4.1}. Experimental results reveal that our Confidence-aware Structure Learning (CSL) achieves consistent performance improvements across architectures, demonstrating its effectiveness as an architecture-independent framework.

\section{More Analysis for residual dispersion}
\label{sec:residual}
\subsection{Why residual dispersion but not other common metrics}

The choice of residual dispersion as a reliability metric stems from its theoretical foundation in entropy minimization principles. As derived in \cref{eq:entropy_decomp}-\cref{eq:v_def}, the cross-entropy objective naturally decomposes into two complementary terms: the maximum confidence $p_n(k')$ and residual dispersion $v_n$. This decomposition reveals an intrinsic geometric relationship, that reliable predictions must simultaneously maximize confidence in the dominant class and dispersion among residual probabilities.

Traditional metrics like entropy $H(p_n)$ and prediction margin $m_n$ fail to meet this criterion. Though entropy theoretically encourages unimodal distributions, it inadvertently tolerates pathological multi-peaked configurations. Consider predictions $p_A = [0.5, 0.5, 0,\ldots,0]$ and $p_B = [0.5, 0.01,\ldots,0.01]$. Paradoxically, $p_A$ exhibits lower entropy despite being less reliable. This demonstrates entropy inability to distinguish valid unimodal predictions from problematic multi-modal ones.

Residual entropy $H_{\text{res}} = -\sum_{k \neq k'} p_n(k)\log p_n(k)$ avoid such problem and is ostensibly similar to the second term in \cref{eq:entropy_decomp}, but its additional linear dependence on $pn_(k)$ significantly reduces its ability to judge prediction credibility under overconfidence.

Margin $m_n = p_n(k') - \max_{k \neq k'} p_n(k)$ focuses only on the top two categories. In the case of overconfidence, the margin is completely dominated by the maximum confidence and shows no discrimination.

\begin{table}[t]
  \centering
  \begin{tabularx}{\columnwidth}{*{4}{>{\centering\arraybackslash}X}|*{1}{>{\centering\arraybackslash}X}|*{1}{>{\centering\arraybackslash}X}}
    \toprule[1.1pt]
     $v_n$     & $H(p_n)$  &  $H_{res}$ & $m_n$      & 1/8    & 1/4 \\
    \midrule
    \checkmark &           &            &            & \textbf{79.6}&\textbf{80.3}\\
    \checkmark &\checkmark &            &            & 77.3 & 78.5 \\
    \checkmark &           &\checkmark  &            & 77.5 & 79.1 \\
    \checkmark &           &            & \checkmark & 76.9 & 78.2 \\
    \checkmark & \checkmark& \checkmark & \checkmark & 71.5 & 76.3 \\
    \bottomrule[1.1pt]
  \end{tabularx}
  \caption{Comparison of different combinations of metrics, evaluated on the original PASCAL VOC 2012 with a crop size of 321.}
  \label{tab:11}
\end{table}
\subsection{Why not use multiple metrics}

While combining multiple metrics theoretically enhances pseudo-label selection with negligible additional time overhead, considering that metrics like entropy or margin are repeated measurements of \cref{{eq:entropy_decomp}}, adding such redundant features will introduce covariance conflicts that will degrade the model performance. As shown in \cref{tab:11}, optimal performance is achieved when only maximum confidence and residual dispersion are considered.

\section{Potential Limitations}
\label{sec:limit}
In CSL, we employ a convex optimization strategy within the confidence distribution feature space to exclude potential high-confidence erroneous pseudo-labels caused by model overconfidence. However, we observe that when the proportion of labeled data is extremely small, the significant disparity in the marginal distributions between sample sets leads to unavoidable cognitive bias. This limitation is particularly pronounced in real-world scenarios, where such imbalanced data splits are common. Thus, an important avenue for future research could involve leveraging limited labeled data more effectively to calibrate cognitive biases and improve the quality of pseudo-labels.

Additionally, although the theoretical validity of residual dispersion is established under the general principles of semi-supervised semantic segmentation, given the broad applicability of entropy minimization, this proof can be readily extended to similar domains such as semi-supervised classification or unsupervised domain adaptation. Similarly, introducing direct confidence calibration methods commonly used in other fields into semi-supervised semantic segmentation represents another promising technical pathway. We leave these potential extensions for future exploration.

Last but not least, CSL utilizes the masking of reliable regions to leverage contextual relationships, thereby enhancing the model's learning in low-confidence areas. However, we find that as the interfering information surrounding low-confidence regions is masked, predictions in these areas tend to be more accurate compared to unmasked regions. Yet, due to the potential for errors, the information from these areas is discarded during training. Therefore, a potential approach could be to further screen this valid information. This could potentially complement the direct supervision signals for low-confidence regions, fostering a more effective learning process.

\section{More Visualizations}
\label{sec:visual}
\begin{figure*}[t]
  \centering
   \includegraphics[width=1\linewidth]{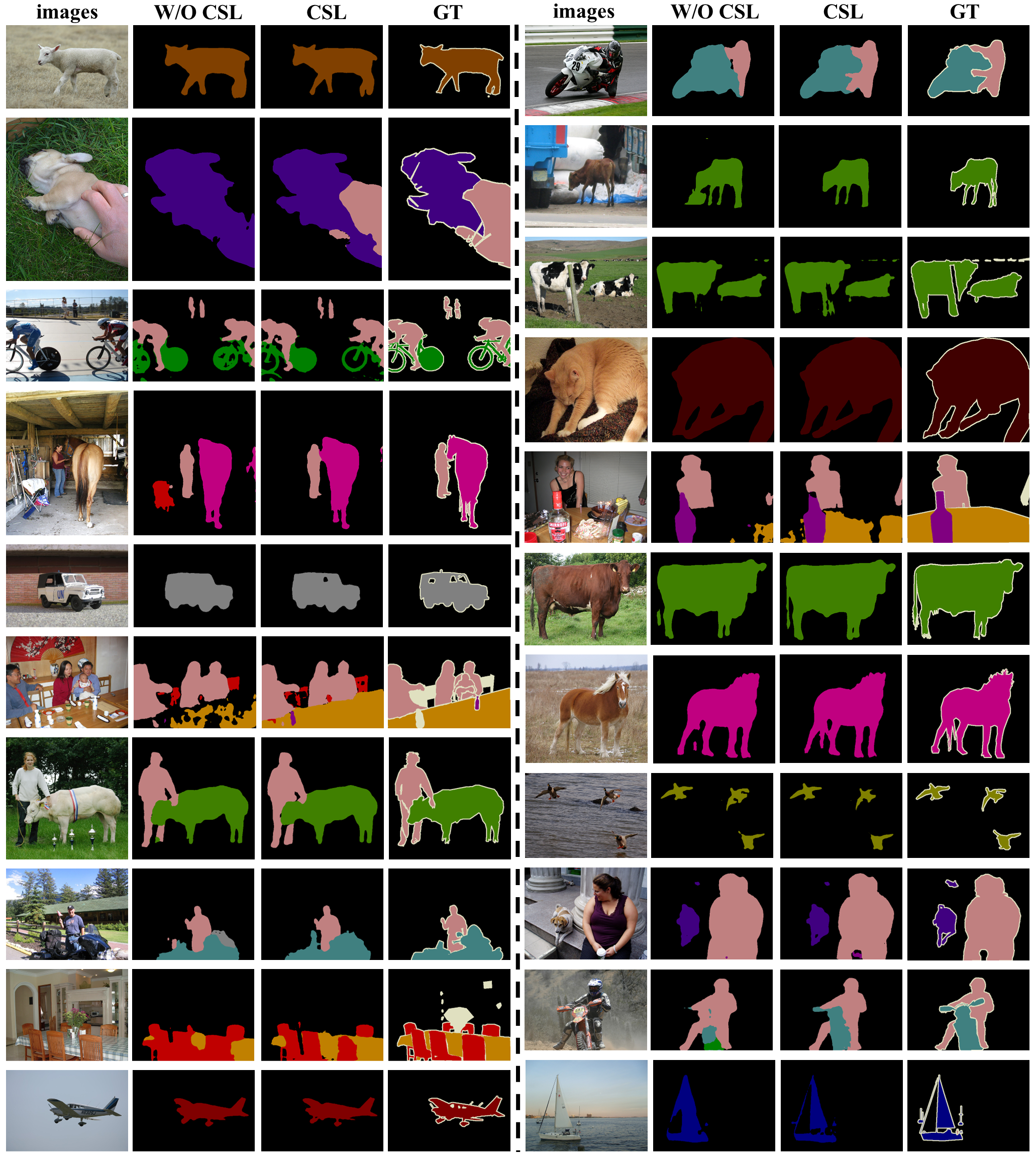}
   \caption{More visualization of the segmentation results on the PASCAL VOC 2012 dataset on  1/8 splits with a crop size of 513.} 
   \label{fig:8}
\end{figure*}

In \cref{fig:8}, we show that the method in this paper and other methods More segmentation results on the PASCAL VOC 2012 dataset.

\end{document}